\documentclass{article}
\usepackage{graphicx, array, multirow} % Required for inserting images
\usepackage{xcolor}
\usepackage[normalem]{ulem}
\usepackage{amsmath}
\usepackage{amsfonts}
\newcommand\norm[1]{\left\lVert#1\right\rVert}
\newcommand{\code}{\texttt}
\usepackage{algorithm}
\usepackage{algpseudocode}
\usepackage{fancyhdr}

\title{Multi-Texture Synthesis through
Signal Responsive Neural Cellular Automata}
\author{Mirela-Magdalena CATRINA\footnote{Faculty of Mathematics and Informatics, {\it Transilvania} University of Bra\c sov,  Romania, e-mail: mirela.catrina@student.unitbv.ro}, \\ Ioana Cristina PLAJER\footnote{Department of Mathematics and Informatics, {\it Transilvania} University of Bra\c sov,  Romania, e-mail: ioana.plajer@unitbv.ro}, Alexandra B\u AICOIANU\footnote{Department of Mathematics and Informatics, {\it Transilvania} University of Bra\c sov,  Romania, e-mail: a.baicoianu@unitbv.ro}}
\date{}

\begin{document}

\maketitle

\section*{Abstract}
Neural Cellular Automata (NCA) have proven to be effective in a variety of fields, with numerous biologically inspired applications. One of the fields, in which NCAs perform well is the generation of textures, modelling global patterns from local interactions governed by uniform and coherent rules. This paper aims to enhance the usability of NCAs in texture synthesis by addressing a shortcoming of current NCA architectures for texture generation, which requires separately trained NCA for each individual texture. In this work, we train a single NCA for the evolution of multiple textures, based on individual examples. Our solution provides texture information in the state of each cell, in the form of an internally coded genomic signal, which enables the NCA to generate the expected texture. Such a neural cellular automaton not only maintains its regenerative capability but also allows for interpolation between learned textures and supports grafting techniques. This demonstrates the ability to edit generated textures and the potential for them to merge and coexist within the same automaton. We also address questions related to the influence of the genomic information and the cost function on the evolution of the NCA.

\section{Introduction}

Texturing is a demanding, complex, and fundamental process in computer graphics \cite{SOTA}, referring to the process of mapping a texture, usually provided by an image file, onto a given object. The technique of producing high-quality textures of custom size, which are similar to the given examples without containing artifacts or unnatural repetition is called texture synthesis. The importance of this technique is rapidly increasing due to its extensive use in industries like game and film production, combined with the labor-intensive process of manual texture creation.

Texture synthesis by procedural generation offers multiple advantages, such as efficient sampling and use of time and memory.  A few lines of shader code can create complex, satisfactory patterns and improve the system efficiency compared to sampling from 2D image textures, generated using example-based methods \cite{SOTA, mordvintsev:uNCA, mordvintsev:Self-organisingTex}. These parametric methods, varying in diversity and size, provide minimal errors and are highly scalable to any resolution output. It is not unexpected that neural networks have been shown to be a powerful approach to texture synthesis \cite{gatys:TexSynthCNN, ulyanov, xian:texGAN}.

Recent work demonstrates the capabilities of neural cellular automata (NCA) for this task and highlights the advantages of using this approach, namely, efficient sampling, compactness of the underlying representations and easy usage through short shader programs \cite{mordvintsev:uNCA, mordvintsev:Self-organisingTex, pajouheshgar2022dynca, xu:TexSynthGNCA}. The NCA is employed as a differentiable image parametrization \cite{mordvintsev:Self-organisingTex, mordvintsev:DifferentiableImgParam} meaning it transforms a given image, which is initially, a gray uniform image, in the style of the given target or example. It is important to underline, that the automaton does not create a pixel-copy of the target image, but it generates an example-based texture. The NCA works as a generator and uses the gradients provided by a pre-trained differentiable model, in order to learn the style of the target. This means that each trained NCA generates one single targeted texture and, consequently, such an approach for large-scale multiple texture generation may become burdensome.

Nevertheless, after multiple experiments and research, we concluded that NCAs could perfectly fit the task of generating textures, given the fact that they model global patterns from local interactions governed by uniform, consistent rules, enhancing structured, near-regular patterns in a compact manner. Therefore, by this study we aim to enhance the usability of NCAs in texture synthesis by training a single neural cellular automaton to evolve into multiple textures, thus increasing the automaton's generalisation capacity. Our solution relies on providing texture information into the seed, as a genomically-coded internal signal that the NCA will interpret and thus yield the expected texture. Particularly, we define a few hidden channels of the cell's state to be genome channels, and use binary encoding for each example image index. As a result our NCA will be able to develop 2, 4, 8 (or any other power of two) number of textures. We detail the process of selecting an optimal baseline for our architecture in Section \ref{sec:2.1SingleTex} and cover the implementation details of the proposed approach in Section \ref{sec:2.2MultiTex}. The results, consisting of an NCA that evolves up to 8 textures, are showcased in Section \ref{sec:3Results}. Moreover, considering the increased complexity and extended range of behaviours exhibited by the NCA, we explore interpolation behaviour and grafting techniques, further emphasizing the editability of our generated textures and the possibility of multiple textures joining and coexisting in one single automaton. In Section \ref{sec:4DiscusAnalysis}, we also address questions relating to the extent to which our NCA uses the genomic information, and proper loss function selection based on the properties of selected example images.

\section{Methodology}
\label{sec:2Methodology}

In this section we introduce the methodology and the theoretical aspects important to our scope. Firstly we introduce the architecture of an NCA for single texture generation and we describe the way it evolves in a given texture, specifically focusing on the loss function and cell perception methodologies.

In the following, we explain the NCA model for multiple texture generation, which is the main scope of our work. The experiments follow the proposed genomic coding in the seed for the development of multiple textures. We explain the reasoning for choosing this type of encoding, as well as the architecture and the training details. Furthermore, we briefly discuss the new behaviour of texture interpolation specific to our NCA and its potential.

Finally, in this section, we introduce further capabilities of the proposed architecture, like regeneration and grafting. Clear and concise steps are presented for these experiments and the results are detailed in Section \ref{sec:4DiscusAnalysis}. 

\subsection{Single texture generation NCA}
\label{sec:2.1SingleTex}

Starting from the architecture discussed in \cite{mordvintsev:Self-organisingTex} this study aims to determine an optimal baseline for further expanding the NCA for multi-texture generation. Our experiments included choosing optimal perception kernels, an adequate neural network architecture and an appropriate loss function.

\subsubsection{Architecture and Inference}

A neural cellular automaton is a model that encompasses cells displayed in a structured manner. Different cellular automaton types were developed in literature, each designed for a specific use-case \cite{otherUse:LandUseNCA, cavouti:signalsAdversarial, otherUse:NCALevelGen,  softRobotsRegen, mordvintsev2020growing}. 

For the texture synthesis task, alive cells are displayed in a grid and each one corresponds to a pixel in the generated texture. 
The NCA is initialized with a seed state, $s_0$, and is guided through time using an update function applied for each cell. At each timestamp, cell states are modified, and thus the expected texture emerges after a few evolution stages. The update rule is learnt by a neural network and outputs the new state of a cell based on its previous state and those of its neighbours.

Each cell's information is stored in an $n_s$-dimensional state vector, where $n_s$ is a hyperparameter of the model. The state vector for each of the cells comprises two components: 
\begin{itemize}
    \item \textbf{3 color channels}: the first 3 values of the state vector correlate with the RGB channels, assigning color to the corresponding pixel
    \item \textbf{$n_h$ hidden channels}: the following values or channels, designed to facilitate cell communication ($n_h = n_s-3$). orangeIn our experiments, we preponderantly used $n_h \in \{9, 10, 12\}$ hidden channels, specific per-experiment values are detailed in Section \ref{sec:3Results}.
\end{itemize}

\begin{figure}[]
    \centering
    \includegraphics[width=350px]{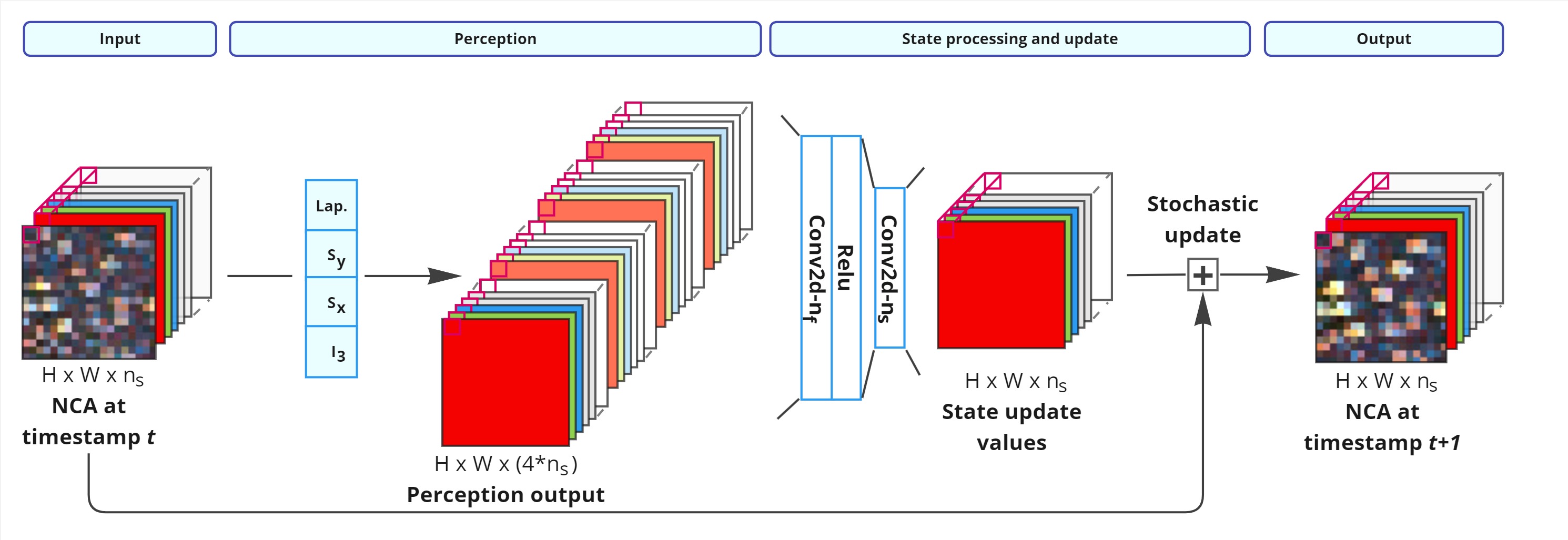}
    \caption{One NCA pass.}
    \label{fig:nca-pass}
\end{figure}

For single texture generation we use a NCA based on the architecture proposed in \cite{mordvintsev:Self-organisingTex}. A NCA pass, applied at the transition from time step $t$ to $t+1$, is illustrated in Figure \ref{fig:nca-pass} and consists of two major steps: perception, and state processing and update. The perception stage employs 4 fixed kernels: $I_3$ - $3 \times 3$ identity kernel, $S_x$ - $Sobel_x$, $S_y$ - $Sobel_y$, $Lap$ - $Laplacian$ to convolve with each channel of the NCA state and offers the perception values for the next stage. We used the 9-point variant of the discrete Laplace operator, as recommended in \cite{mordvintsev:Self-organisingTex}. The kernel values for $S_x, S_y, Lap$ are displayed in Equation \ref{eq:SobelLaplaceKernels}. 

\begin{equation}
     S_x = \left[\begin{array}{rrr}
            -1& 0 & 1 \\
            -2& 0 & 2 \\
            -1& 0 & 1 
        \end{array}\right];S_y = \left[\begin{array}{rrr}
            -1&-2 &-1 \\
            0 & 0 & 0 \\
            1 & 2 & 1 
        \end{array}\right]; Lap = \left[\begin{array}{rrr}
            1 & 2 & 1 \\
            2 &-12& 2 \\
            1 & 2 & 1 
       \end{array}\right]
    \label{eq:SobelLaplaceKernels}
\end{equation}

Different kernels have been studied and are discussed in further sections. Also, during this stage circular padding is applied to ensure the preservation of the image size and the visual tileability of the generated texture. The concatenated results provided by the 4 kernels are fed to the neural network that models the update rule as they provide each cell's state together with data from its neighbourhood.

\begin{figure}[]
    \centering
    \includegraphics[width=300px]{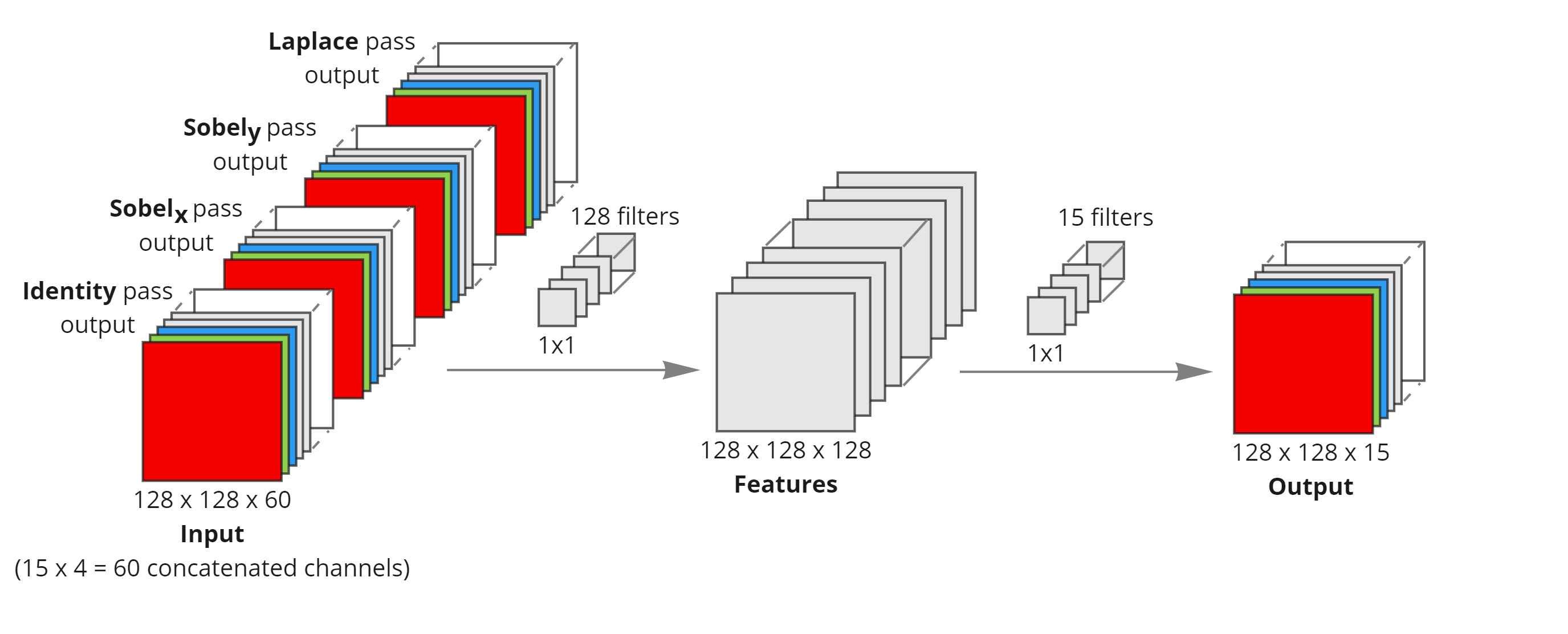}
    \caption{The NN that models the update rule based on the perception output.}
    \label{fig:texnn-structure}
\end{figure}

The neural network employed for learning the update rule is displayed in Figure \ref{fig:texnn-structure}. It is important to note that all of its filter kernels are of size $1 \times 1$, meaning a cell only updates its state based on the input received by the perception, and does not interfere with the processing of other states. This simulates the per-cell, independent, state processing. The first convolutional layer employs $n_f$ filters. In our experiments we use different values for the $n_f$ hyperparameter. These specific values are presented in Section \ref{sec:3Results}. The neural network generates the values needed to update each cell's state, resulting in an output size $h \times w \times n_s$ ($h$ - height, $w$ - width) similar to the input automaton state,  and the necessity of using $n_s$ filters in the second convolutional layer.

The output modification values are applied stochastically, to break symmetry and relax the expectancy of global synchronization for our self-organising system. This can also be seen as a per-cell dropout \cite{mordvintsev:Self-organisingTex}. In our experiments, half the number of the total cells updates at each timestamp.

Using this architecture, we can generate textures of varying sizes. During training, we use $128 \times 128$ examples. At inference, we can generate both $128 \times 128$ textures and smaller or larger rectangular textures.

\subsubsection{Training of the NCA}
\begin{figure}[h!]
    \centering
    \includegraphics[width=300px]{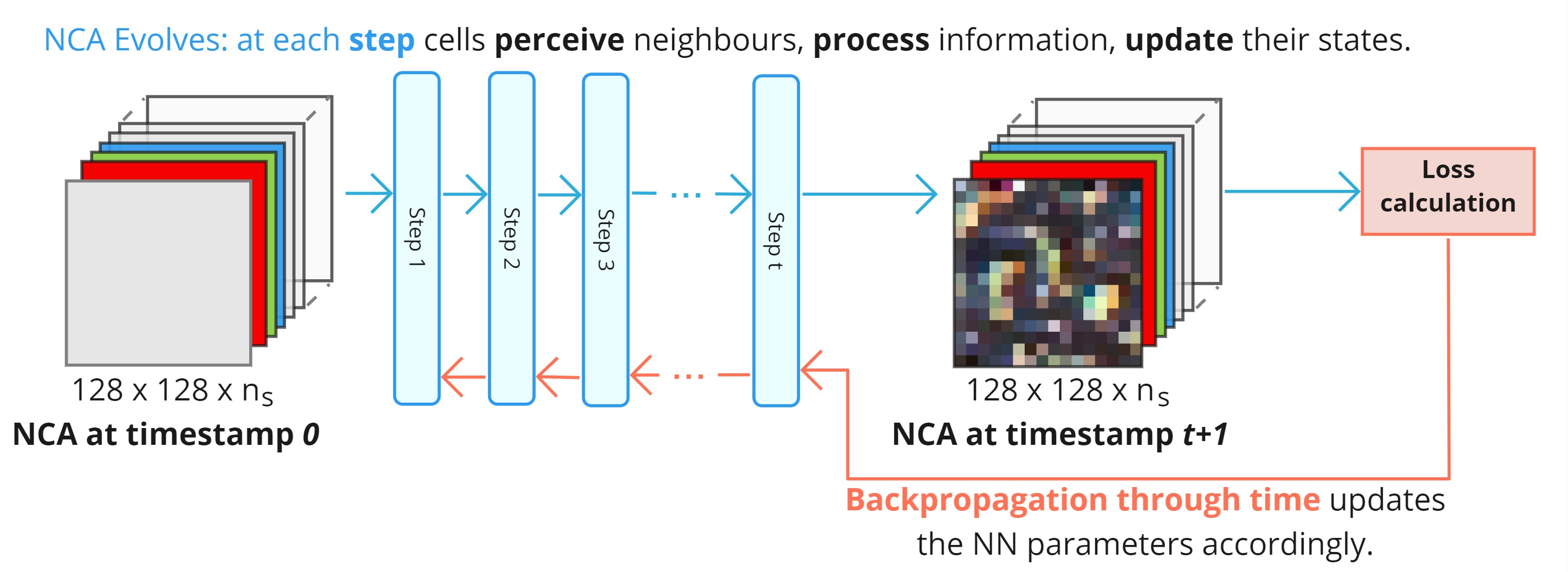}
    \caption{A training step. The NCA runs iteratively through the NN for $t$ steps, as illustrated in Figure \ref{fig:nca-pass}. The loss is then calculated and the the NN's parameters are updated by backpropagation through time.}
    \label{fig:loss-to-backprop-H1}
\end{figure}

In the current context, training the NCA involves training the neural network that models the update rule. We initialize the NCA with a uniform state and evolve it iteratively for $t$ steps. At timestamp $t+1$ we measure the quality of the generated texture against the provided example and apply backpropagation to the neural network. In our experiments, $t$ is a randomly generated number between $32$ and $96$, as utilized and tested in \cite{mordvintsev:Self-organisingTex}. The training proccess is illustrated in Figure \ref{fig:loss-to-backprop-H1}.

\textbf{Loss function}

Our NCA should imitate a given example's style, not generate a pixel copy of it. Therefore, we must train our state processor, represented by the neural network illustrated in Figure \ref{fig:texnn-structure}, accordingly. Most style transfer methods rely on the distributions of feature maps provided by a neural network whose layers are considered to capture style. We name it an \textit{observer} or \textit{differentiable texture discriminator} \cite{mordvintsev:DifferentiableImgParam, mordvintsev:arxivTexGen}. We pass the RGB channels of our NCA and the example image through the observer and drive the selected feature maps distributions to match.

\begin{figure}[!h]
        \centering
        \includegraphics[width=270px]{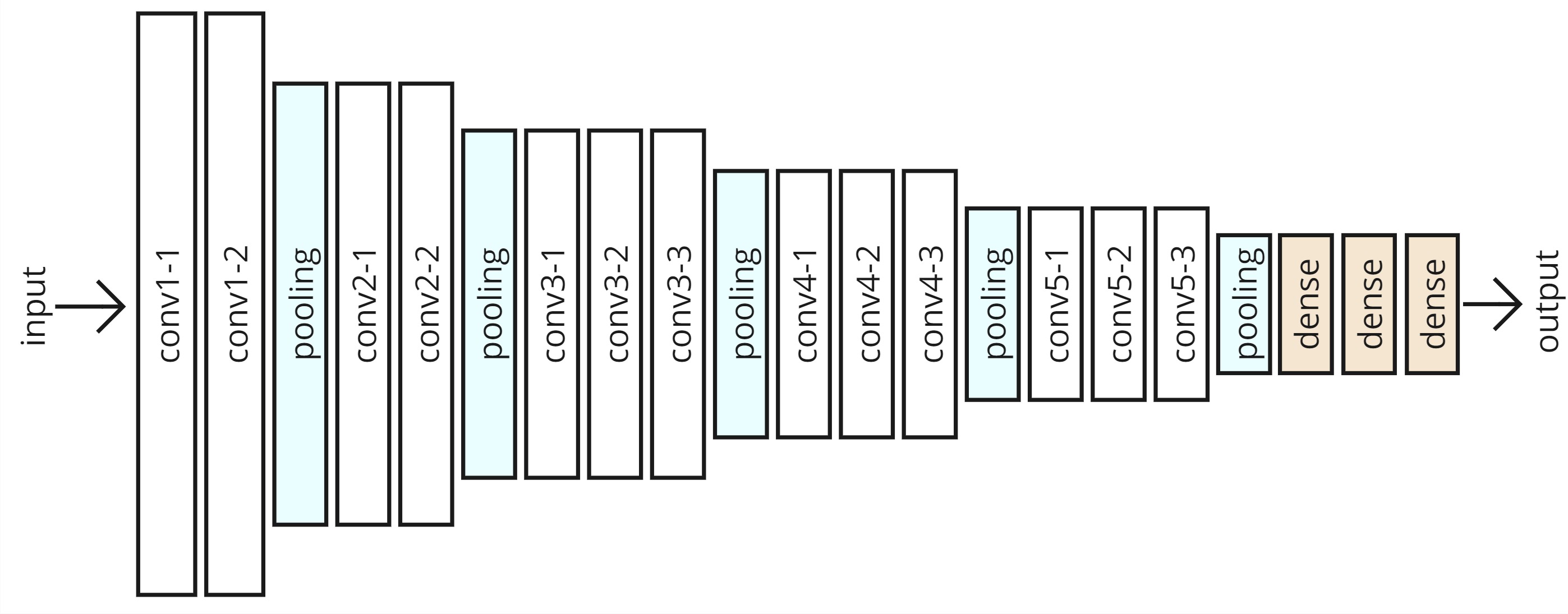}
        \caption{VGG16 architecture}
        \label{fig:vgg16-archi}
    \end{figure}

\begin{figure}[h!]
    \centering
    \includegraphics[width=250px]{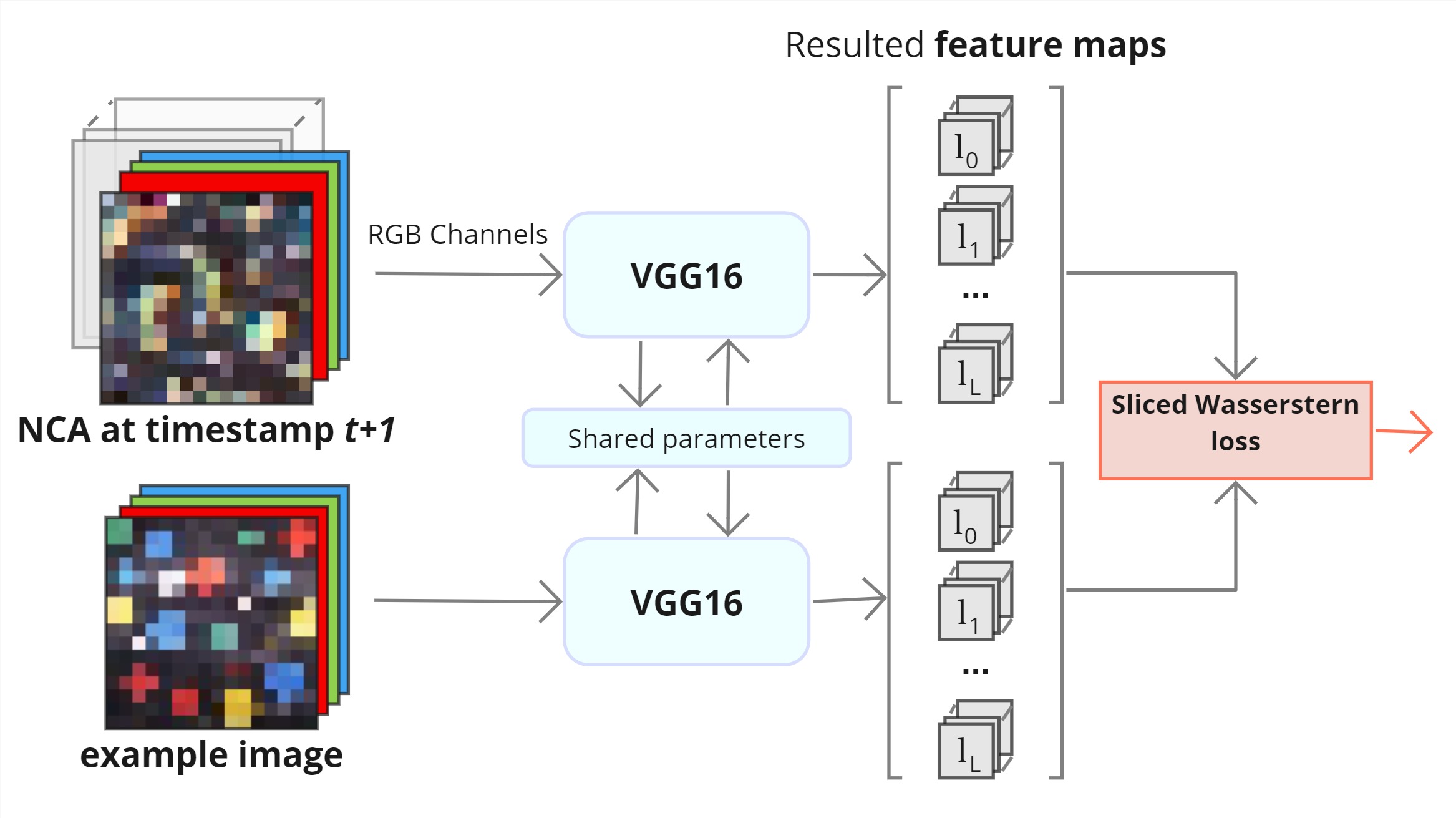}
    \caption{The loss calculation process. The feature distributions for the specified $L$ layers are extracted by passing the state's RGB channels through VGG16. They are then matched to those of the example image using SWL.}
    \label{fig:loss-to-backprop-H2}
\end{figure}

The NCA uses a trained VGG16 network as the differentiable texture discriminator. The choice of VGG is deliberate, as most style transfer approaches utilize VGG variants due to superior results compared to other architectures \cite{mordvintsev:DifferentiableImgParam, mordvintsev:Self-organisingTex}. The activations provided by a selection of $L=5$ layers from the VGG16 network (conv1-1, conv2-1, conv3-1, conv4-1, conv5-1) is used, considering the  architecture as shown in Figure \ref{fig:vgg16-archi}. The loss calculation process is illustrated in Figure \ref{fig:loss-to-backprop-H2}.

Although most texture synthesis algorithms rely on feature distribution matching based on Gram matrices \cite{heitz:OTLoss, mordvintsev:Self-organisingTex}, their limitations have been thoroughly discussed. The recent work of \cite{heitz:OTLoss} demonstrates the superiority of the Sliced Wasserstein Loss (SWL) in capturing the complete set of feature distributions. Therefore, we employ the SW Loss for our experiments. 

We select feature maps from $L$ layers of the observer network for the target image $I$, and the generated image, $\tilde{I}$. We denote the textural loss by $\mathcal{L}(I, \tilde{I}) \in \mathbb{R}^+$. For a layer $l$ we have $M_l = h \times w$ ($w$ - height, $w$ - width) feature vectors: $F_m \in \mathbb{R}^{c_l}, m \in \{0, 1, \dots, M_l-1\}$, where $c_l$ is the depth dimension of the feature map. The set of distributions for these vectors is noted $p^l$ for the features obtained by passing the example image through the VGG, respectively $\tilde{p}^l$, obtained by passing the generated image through the VGG.

The SWL operates on the sets of distributions $p^l$ and $\tilde{p}^l$, for all $l \in L$. Its value is the sum of the distances between the distributions extracted from $I$ and $\tilde{I}$ at each layer $l$, as formalized by 

\begin{equation}
    \mathcal{L}_{SW}(I, \tilde{I}) = \sum_{l=1}^{L}{\mathcal{L}_{SW}(p^l, \tilde{p}^l)}
    \label{eq:SWL_img}
\end{equation}

The distance between the distributions $p^l$ and $\tilde{p}^l$ is calculated by approximating the optimal transport (OT) between them (note that the transport map is not optimal but the optimized distribution is proven to converge towards the target distribution \cite{heitz:OTLoss}): 

\begin{equation}
    \mathcal{L}_{SW}(p^l, \tilde{p}^l) = \mathbb{E}_v[\mathcal{L}_{SW1D}(p_V^l, \tilde{p}_V^l)]
    \label{eq:SWD}
\end{equation}

The approximation is done through the SW distance, as defined in Eq \ref{eq:SWD}: the expectation of the one dimensional optimal transport (1D OT) distances after projecting the feature points onto random directions $V \in \mathcal{S}^{c_l}$ (here, $p_V^l$ and $\tilde{p}_V^l$ correspond to the projections of the feature points on a direction $V$). Projecting over several directions leads our histogram to converge towards the target histogram \cite{pitie:OTLoss}.

We project $p^l$ and $\tilde{p}^l$ feature vectors on a random direction using dot product with the vector $V$ and obtain two sets of scalars. The 1D OT is calculated by sorting the scalars and applying Mean Squared Error (MSE), known also as $L_2$ norm, over the obtained sets:

\begin{equation}
    \mathcal{L}_{SW1D}(S, \tilde{S}^l) = \frac{1}{|S|} \norm{sort(S) - sort(\tilde{S})}^2
    \label{eq:OT-dist}
\end{equation}

\begin{figure}[]
    \centering
    \includegraphics[width=300px]{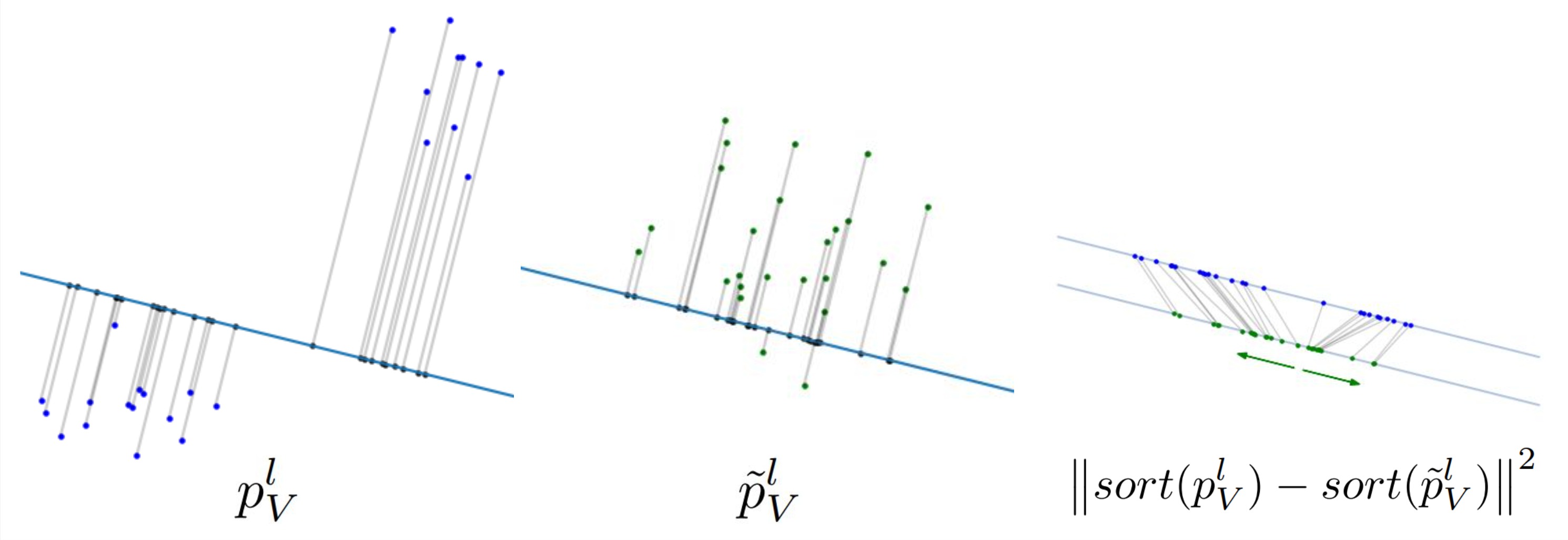}
    \caption{The Sliced Wasserstein loss visualized for a 2D example. By computing the $L_2$ difference between sorted lists of projections from the target distribution ($p_V^l$) and the obtained distribution ($\tilde{p}_V^l$) we push the NN to reach the targeted distribution. This process is done for several projection directions to capture all necessary information.}
    \label{fig:SWL-explanation}
\end{figure}

A simplified visualization of the SWL is illustrated in Figure \ref{fig:SWL-explanation} and a pseudocode for this algorithm is presented in Algorithm \ref{alg:SWD-loss}. First, we reshape the feature tensor of shape $h \times w \times c_l$ into $M_l \times c_l$. We then project these features over 32 random directions (unit vectors of dimension $c_l$), sort them and measure the $L_2$ distance. The strategy of applying this loss is portrayed in Figure \ref{fig:loss-to-backprop-H2}. A detail not covered in the figure for simplicity is the overflow loss term, given in equation \ref{eq:LOverflow}, that we add alongside the SWL to keep all state channels in the interval $[-1, 1]$.

\begin{equation}
    \mathcal{L}_{overflow}(I) = \sum_{x \in I}{|x - clip_{[-1,1]}x|} 
    \label{eq:LOverflow}
\end{equation}

Note that $clip_{[-1,1]}x$ returns a vector of the same dimension, with all values clipped in the $[-1, 1]$ interval. This stabilises training by preventing drift in latent channels and aids in post-training quantisation \cite{mordvintsev:uNCA}. The complete formula for our loss is:
\begin{equation}
    \mathcal{L}(I, \tilde{I}) = \mathcal{L}_{SW}(I_{RGB}, \tilde{I}) + \mathcal{L}_{overflow}(I)
\end{equation}

\renewcommand{\algorithmicrequire}{\textbf{Input:}}
\renewcommand{\algorithmicensure}{\textbf{Output:}}
\begin{algorithm}
\caption{Sliced Wasserstein Loss Computing}\label{alg:SWD-loss}
\begin{algorithmic}
\small
\Require \\ $source_l$ - features received from VGG's $l$ layer for the NCA output, with $(c_l,M_l)$ shape \\ $target_l$ - features received by the same layer for the example image of the same shape

\Ensure the SWL for the given layer - $\mathcal{L}_{SW}(p^l, \tilde{p}^l)$
\Function{sw\_loss}{$source_l, target_l$}
\State $proj\_n \gets 32$ \Comment{number of projections}
\State $Vs \gets rand\_like(c_l, proj\_n)$ \Comment{random directions (normalized values)}
\State $source\_proj \gets dot(source_l, Vs)$ \Comment{project source in Vs}
\State $target\_proj \gets dot(target_l, Vs)$ \Comment{project target in Vs}
\State $diff \gets sort(source\_proj, axis=1) - sort(target\_proj, axis=1)$
\State \Return $mean(square(diff))$
\EndFunction
\end{algorithmic}
\end{algorithm}

\textbf{Pooling}

We pass NCA states to the training step in batches. A \textit{sample pool} based strategy is necessary for the long-term stability of the automaton \cite{noi:NCA, mordvintsev2020growing, mordvintsev:Self-organisingTex}, explicitly the behaviour of the NCA outside the few steps considered during training. If no pooling strategy is applied, the behaviour of the automaton after the 96 steps threshold would be unstable. We employ the same strategy presented in \cite{mordvintsev2020growing, mordvintsev:Self-organisingTex} meaning we construct a pool of 1024 future textures, and seed states of the NCA. Batches of 8 elements are selected from this pool, and after the training step presented in Figure \ref{fig:loss-to-backprop-H1} the new states are placed back into the pool. Also, at batch selection, the highest loss scoring element is replaced with a seed - the uniformly initialized image - to enforce seed evolution, and texture stability and avoid training on irrelevant hallucinations during the first stages of training. 

\subsection{Multi-texture generation}
\label{sec:2.2MultiTex}

Signals are the central component of animal interaction and an essential part of animal life. Internal signals include hormonal or neural signals within the body, bioelectrical and genomic signals during development \cite{Stovold:NCARespondSignals}.  Research studying the integration of internal signals in the morphogenesis model of NCA has been conducted \cite{cavouti:signalsAdversarial, Stovold:NCARespondSignals, kuriyama:signalsGradientNCA}, but not in the context of self-organising textures.

\begin{figure}[!h]
    \centering
    \includegraphics[width=200px]{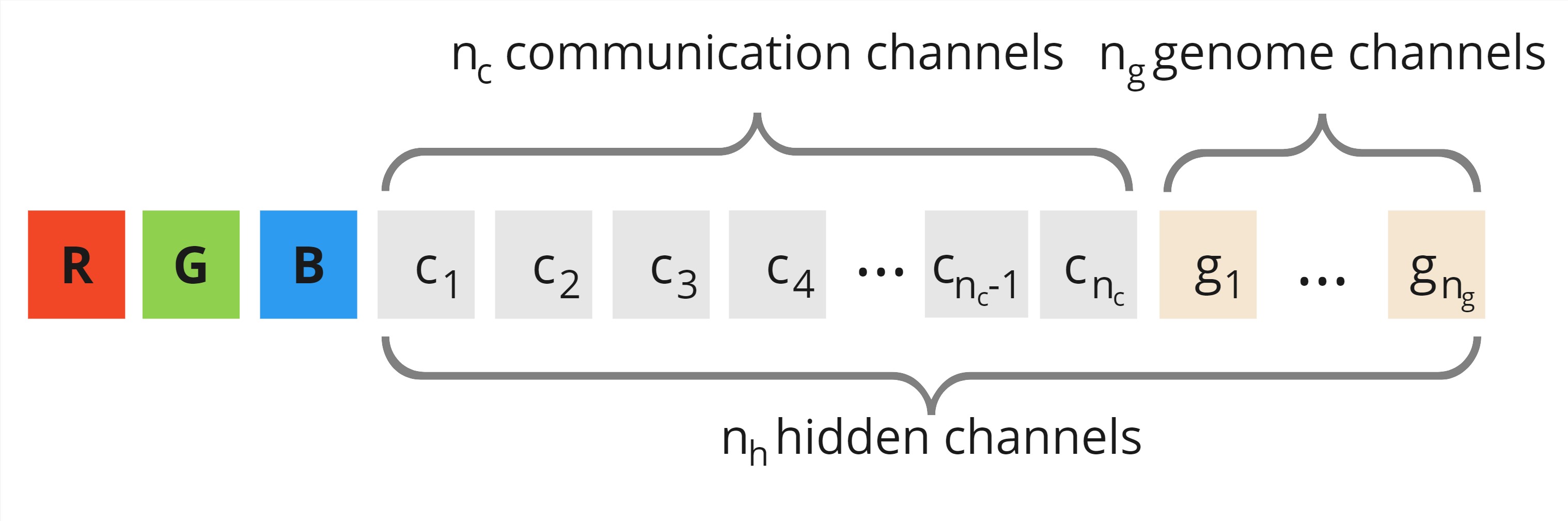}
    \caption{The state vector of a cell in a multi-texture generation NCA. It comprises $n_s$ components: 3 RGB channels, $n_h$ hidden channels consisting of $n_c$ communication channels and $n_g$ genome channels. %These hyperparameters are selected per experiment and detailed below.
    }
    \label{fig:genomic-seed-state}
\end{figure}

Integrating internal signals in our NCA boils down to attributing special meaning to a few channels of the cell's hidden state. As illustrated in Figure \ref{fig:genomic-seed-state}, the hidden $n_h$ channels of the hidden state are now divided into $n_c$ communication channels and $n_g$ genomic channels.

\begin{figure}[]
    \centering
    \includegraphics[width=300px]{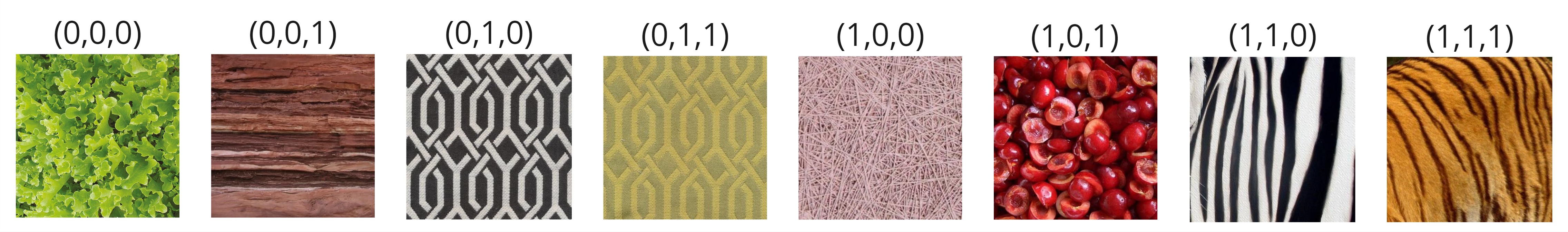}
    \caption{Examples of textures generated by a 3-genome NCA architecture. The tuple for each texture corresponds to the $(g_1, g_2, g_3)$ channels in the seed state. This means that, for example, if we want to create a cherry texture, at timestamp 0 all cells would have the state vector with all values set to 0, except the last and second-to-last position which would be set to 1.}
    \label{fig:genome-examples}
\end{figure}

Our experiments include $n_g=1$, $n_g=2$ and, respectively, $n_g=3$ genomic channels, corresponding to 2, 4 and 8 different textures generated by one NCA. Since we employ the overflow loss to stabilise training, encouraging all channels to hold values in the [-1, 1] interval, we needed an appropriate encoding strategy. Therefore we chose binary encodings, as genome channel values will initially be either 0 or 1. Furthermore, $n_g$ genomic channels allow generation of $2^{n_g}$ textures and a texture index in its binary coding will allow (ideally) interpolation with all other textures, an advantage further detailed in Section \ref{sec:2.3OtherExp}. An example further discussed in the results section is depicted in Figure \ref{fig:genome-examples} where we see the correspondence between the expected textures and the genomic encoding. The genome channels are set at timestamp 0 for each cell in the automaton as follows: for a frilly texture, we set them to 000, for a stratified texture - 001 and so on. In pseudocode, this translates to the function written in Algorithm \ref{alg:mtg-initialization}. In the algorithm, we initialize all cells in the automaton to be of genome $g$, by setting their last $n_g$ channels ($seed[:,:,-n_g:]$) to the binary encoding of the genome index $g$. All other values are 0.

\begin{algorithm*}
\caption{Automaton genome-based initialization }\label{alg:mtg-initialization}
\small
\begin{algorithmic}
\Require \\
    $h, w$ = height and width of cell plane (size of texture) \\
    $n_h$ = number of cells' state hidden channels, with $n_g$ genome channels, \\
    $g$ =  expected genome index
\Ensure One NCA state of the specified genome
\Function{seed\_of\_genome}{h, w, g}
\State $seed \gets zeros(h, w, 3 + n_h)$
\State $g_{b_2} \gets to\_base\_2(g, n_g)$
\Comment{binary representation of $g$ on $n_g$ bits}
\State $seed [:, :, -n_g:] \gets g_{b_2}$
\State \Return $seed$
\EndFunction
\end{algorithmic}
\end{algorithm*}

After timestamp 1, we do not interfere with any channels: we do not restrict their modifications or values other than encouraging low values through the overflow loss. It is up to the automaton to \textit{understand} and keep the information throughout evolution in order to reach the expected specimen. Discussions on how the automaton understands and preserves the genome details throughout evolution are addressed in the results and discussions section, see Section \ref{sec:4DiscusAnalysis}.

\begin{algorithm*}
\caption{Interpolated genome initialization }\label{alg:interp-initialization}
\small
\begin{algorithmic}
\Require \\
    $n_s$ = number of cell's state channels, with $n_g$ genome channels, \\
    $g_1$ =  first genome index \\
    $g_2$ = second genome index
\Ensure A NCA cell state situated halfway between genomes $g_1$ and $g_2$
\Function{interp\_genome}{$g_1, g_2$}
\State $g1_{b_2} \gets to\_base\_2(g_1, n_g)$
\State $g2_{b_2} \gets to\_base\_2(g_2, n_g)$
\State $g_{result} \gets (g1_{b_2} + g2_{b_2}) / 2$
\State $seed \gets zeros(n_s)$
\State $seed[-n_g:] = g_{result}$
\State \Return $seed$
\EndFunction
\end{algorithmic}
\end{algorithm*}

Apart from generating textures based on the given 2, 4 or 8 example images, we also study the interpolation behaviour for these NCA. Interpolation refers to creating blended textures, a smooth transition between two given examples. Meaning, if the automaton learnt textures based on Figure \ref{fig:genome-examples}, we could also create a mossy wood (combination between the first and second images, starting with a seed with genomic channels set to, for example, g=(0,0,0.5)) or grids on the spectrum defined from the third and fourth textures. Technically, we could interpolate between any two learnt genomes: we interpolate between a genomic code $g_1 = (a_0, a_1, \dots, a_{n_g})$ and $g_2 = (b_0, b_1, \dots, b_{n_g})$ by selecting all pairs where $a_i = b_i$ and setting $c_i = a_i$ and for all pairs where $a_i \neq b_i$ setting $c_i$ to an intermediate value (eg. 0.25, 0.5, 0.75, etc.). The result will be $g_3 = (c_0, c_1, \dots, c_{n_g}).$ Experiments showed that interpolations, where more than one intermediate value has to be set in the genome, are more unstable and unpredictable. The algorithm for creating a genome situated at an equal distance between two learnt genomes is depicted in Algorithm \ref{alg:interp-initialization}. While interpolation behaviour has been studied for multiple texture synthesis models \cite{gotex:65kparam}, it is specific to our architecture in the NCA realm given that other NCAs only develop one texture. Interpolation is performed during inference and does not influence the training strategy.

The optimal adapted training strategy for multi-texture generation relies on a pool with equally divided seeds for each texture. When replacing the highest-scoring state during batch selection, we replace it with a seed with the same initial genome coding. Moreover, we cycle through genome coding replacements: we don't choose the highest-scoring state, but the highest-scoring state belonging to the genome $g_r$, and then move on to the next $g_r$. Practically, we replace from the first batch the highest-scoring state belonging to the genome $g_0$, in the second batch we replace the state belonging to genome $g_1$ and so on. This ensures that more intricate textures, that are harder to learn and yield an overall higher loss during training are not always replaced, since it would lead to instability during the inference for those textures.

Algorithm \ref{alg:pool-mtg} outlines the pooling-based training strategy for generating multiple textures. First, the pool is initialized ($init\_pool(pool\_size)$) by creating $pool\_size/n\_genomes$ seeds of each genome and placing all of them in the shared pool. For each element in the pool, we track the NCA state ($x$) and the index of the corresponding genome ($y$). This tracking is essential because the genome channels in the NCA state change throughout evolution, and we need to identify the corresponding genome at any time for loss calculation and eventual replacement. Next, we initialize the NCA parameters, particularly those of the neural network that models the update rule. The training process runs for a specified number of epochs, where, in each iteration, a batch is selected, prepared, run through the NCA for updates, and then returned to the pool. The NCA weights are adjusted based on the loss calculated for each batch. This iterative process allows the algorithm to evolve and refine the textures within the pool.

\begin{algorithm*}
\caption{Pooling strategy based training for multi-texture generation}\label{alg:pool-mtg}
\small
\begin{algorithmic}
\Require The $pool\_size$ of $nca$ states, $number\_of\_epochs$ and $batch\_size$

\Ensure The trained $nca$, the final pool
\State $pool \gets init\_pool(pool\_size)$
\State $nca \gets init\_nca\_params()$
\For{$iteration \in range(0,number\_of\_epochs)$}

\State $batch \gets \textnormal{random pick }batch\_size \textnormal{ elements from } pool$
\\ \Comment{states and corresponding genome indices}
\State $g_r \gets iteration \% n\_genomes$ 
\If{exists $g_r$ genome in $batch$}
\State $worst_{g_r} \gets state\_with\_highest\_loss\_of\_genome(batch, g_r)$
\Else
\State $worst_{g_r} \gets state\_with\_highest\_los(batch)$
\EndIf
\State $batch[worst_{g_r}].x \gets seed\_of\_genome(batch[worst_{g_r}].y)$
\State $num\_steps \gets random(64, 90)$
\For {$i \in range(0,num\_steps)$}
    \State $batch.x = nca(batch.x)$
\EndFor
\State $loss \gets compute\_loss(batch.x, target\_images(batch.y))$
\State apply $nca$ weights backpropagation
\State $pool \gets \textnormal{put updated individuals of $batch$ back in the pool}$

\EndFor
\State \Return $nca$
\end{algorithmic}
\end{algorithm*}

\subsection{Regeneration and Grafting}
\label{sec:2.3OtherExp}

A well-studied property of NCAs is regeneration \cite{mordvintsev2020growing}. No adaptation of the training strategy is required for the single-texture experiments, and the automaton inherently regenerates and stabilises the pattern shortly after being damaged. However, slight alterations in the training strategy must be applied for the multi-texture architecture. Without these modifications, the NCA generates patches of different textures instead of regenerating the governing texture.

Fortunately, only a few adjustments must be made. The automaton reaches the desired behaviour using the adaptation presented in \cite{mordvintsev:DifferentiableImgParam}. During batch sampling from the pool, besides replacing a high-scoring state we also damage the lowest-scoring state. Damaging here refers to randomizing the state vectors for cells contained in a circle of a radius of 15 to 25 pixels. The adapted training procedure is concisely shown in Algorithm \ref{alg:regen-mtg}, as an altered version of Algorithm \ref{alg:pool-mtg}.

\begin{algorithm*}
\caption{Regeneration-adapted training for multi-texture architecture}\label{alg:regen-mtg}
\small
\begin{algorithmic}
\Require the $pool\_size$ of $nca$ states, $number\_of\_epochs$ and $batch\_size$

\Ensure The trained $nca$, the final pool
\State $pool \gets init\_pool(pool\_size)$
\State $nca \gets init\_nca\_params()$
\For{$iteration \in range(0,number\_of\_epochs)$}

\State $batch \gets \textnormal{random pick }batch\_size \textnormal{ elements from } pool$
\\ \Comment{states and corresponding genome indices}

\State $g_r \gets iteration \% n\_genomes$ 
\State $replace\_highest\_loss\_of\_genome(batch, g_r)$ \Comment{as in Alg. \ref{alg:pool-mtg}}
\State $x_1, x_2 \gets lowest\_2\_losses(batch)$
\State $damage\_texture\_from\_batch(batch[x_1])$
\State $damage\_texture\_from\_batch(batch[x_2])$

\State $nca\_train\_step(nca)$ \Comment{as in Alg. \ref{alg:pool-mtg}}
\State $loss \gets compute\_loss(batch.x, target\_images(batch.y))$
\State apply $nca$ weights backpropagation
\State $pool \gets \textnormal{put updated individuals of $batch$ back in the pool}$

\EndFor
\State \Return $nca$
\end{algorithmic}
\end{algorithm*}

Other experiments include grafting visualizations. Grafting is the act of joining two organisms to continue their growth together. In our case, we consider grafting two or more types of cells, belonging to different genomes, coexisting in one automaton. We do this by initializing, either at timestamp 0 or at a random timestamp, some cells with the seed values of a different genome. By doing this, we enable two textures to coexist and interact in the same automaton. This yields to interesting results, studied in further sections. Cells of different genomes can be structured in multiple ways, by creating concentric circles, stripes etc. of different genome cells. An example for generating a texture where the left half belongs to one genome and the right half belongs to another genome is depicted in Algorithm \ref{alg:grafting-ingerence}.

\begin{algorithm*}
\caption{Grafting inference}\label{alg:grafting-ingerence}
\small
\begin{algorithmic}
\Require $nca$ = nca model with associated hyperparameters \\
$h, w$ = height and width of expected texture \\
$g_0$ = expected genome index for the left half of the texture \\
$g_1$ = expected genome index for the right half of the texture \\
$t\_max$ = timestamp at which to extract the generated texture
\Ensure The grafted texture (RGB image) at timestamp $t\_max$
\Function{graft}{$h, w, g_0, g_1, t\_max$}
\State $left\_seed \gets seed\_of\_genome(h, w/2, g_0)$
\State $right\_seed \gets seed\_of\_genome(h, w/2, g_1)$
\State $texture \gets concat(left\_seed, right\_seed, dim=1)$
\For{$t \in range(0, t\_max)$}
    \State $texture \gets nca(texture)$
\EndFor
\State \Return $texture_{RGB}$
\EndFunction
\end{algorithmic}
\end{algorithm*}

\section{Results and Discussions}
\label{sec:4DiscusAnalysis}
\label{sec:3Results}
%Adaugat ioana
 In order to analyze and evaluate the impact of the different hyperparameters, like size of the first hidden layer and loss function, as well as the influence of the genome on the generation of different texture, a series of experiments were performed. Furthermore the possibility of using the NCA for more complex tasks like interpolation of textures, regeneration of damaged textures and grafting is studied and evaluated. 

The images for all experiments are selected from the Describable Textures Dataset \cite{db:Odtd} that groups of 5640 images, organized in 47 terms (categories) inspired by human perception: banded, bubbly, bumpy, frilly etc. Moreover, we selected a few textures from VisTex Database \cite{db:VisTex}. All experiments were run on T4 GPU and lasted up to 2h each.

\subsection{Results of the experiments}

The single texture generation experiments studied the influence of different perception filters on the quality of the generated texture. For these experiments we used a pool of 1024 images with a batch size of 8. Training was performed for 5000 epochs.% during training.

\begin{figure}[]
    \centering
    \includegraphics[width=160px]{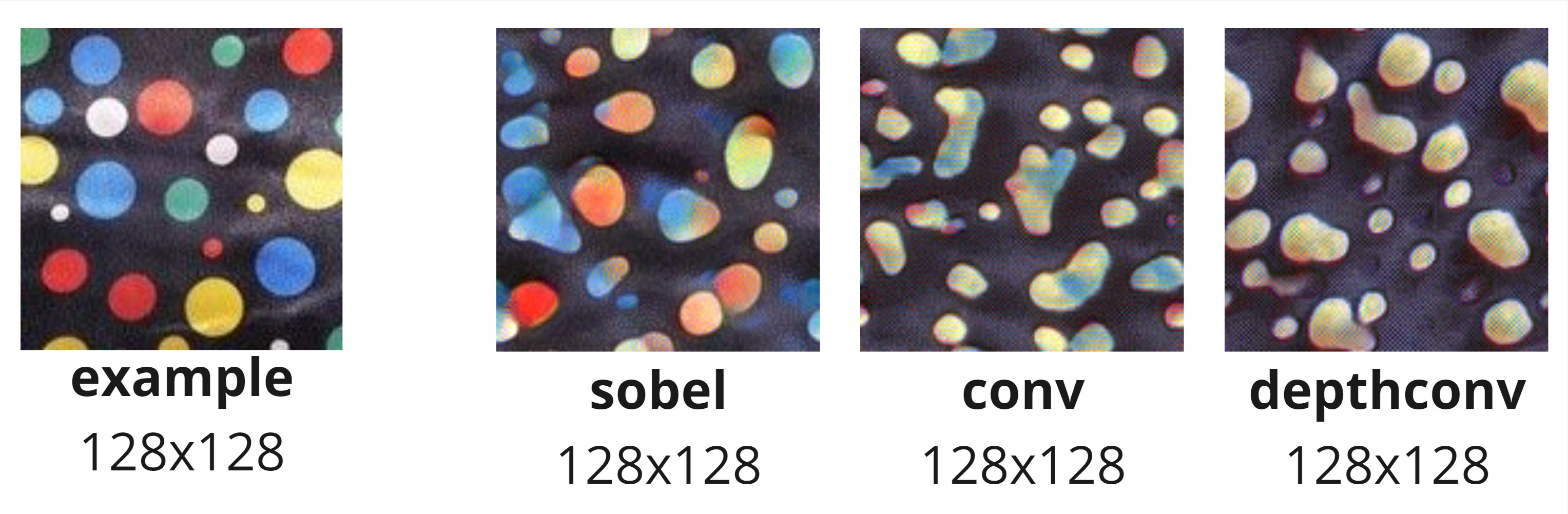}
    \caption{All generation snapshots are taken at timestamp 90. From left to right we see the texture generated with Sobel and Laplace perception kernels, 4 learnt convolution kernels, 4 learnt depthwise convolution kernels.}
    \label{fig:perception-filters}
\end{figure}

As represented in Figure \ref{fig:perception-filters}, the $Sobel_x$, $Sobel_y$, $Laplace$ filters offer enough representation of the neighbors and lead to significantly better results than learned kernels. Of course, leaving the NCA to train longer may enrich the representation provided by learnt kernels. Still, we consider that it would not quantitatively improve the quality output, but it would only add up to the complexity of the NCA representation. %, and that we are satisfied with the results provided by Sobel and Laplacian filters. 
Insight on the influence of image processing filters can be studied in \cite{noi:NCA}, where experiments with other combinations of hardcoded filters have been conducted concerning the influence of such filters on the perception stage.

%Adaugat Ioana
The main focus of this paper is on using a single NCA for the generation of multiple textures, by using the information in the genome channel. Furthermore, some new usages of the NCA were explored. The following experiments and results underline the significance of our work.

\subsubsection{Multi-texture generation and interpolation results}

This subsection details the tackled NCA architectures for multi-texture generation. Specifically, we study different state formats, the long-term stability of generated textures, and address interpolation behaviour between learnt genomes. The NCAs are trained for 10000 epochs, with a pool size of 1024 samples and batch size of 8 images. Table \ref{tab:multitex-architextures} summarizes the performed multi-texture experiments.

\newcommand\T{\rule{0pt}{3ex}}       % Top strut
\newcommand\TS{\rule{0pt}{2.6ex}}       % Top strut
\newcommand\B{\rule[-1.8ex]{0pt}{0pt}} % Bottom strut
\begin{table}[!h]
    \centering
    \begin{tabular}{|c|c|c|c|c|c|}
         \hline
         No. & Experiment & \multicolumn{2}{c|}{$n_h$} & \multirow{2}{*}{$n_f$} & \multirow{2}{*}{Aim} \\
         \cline{3-4}
         genomes & name & $n_c$ & $n_g$ & &\\
         \hline
         \multirow{1}{*}{2} & G2Feasible (G2F)\T\B&9&1&96&feasibility of model \\
         \hline
         \multirow{3}{*}{4} & G4Similar (G4Sim)\T&9&2&128&similar (sim.) textures \\
                            & G4Different (G4Diff)\TS&9&2&128&different (diff.) textures \\
                            & G4Structured (G4Str)\TS\B &9&2&128&structured textures \\
         \hline
         \multirow{3}{*}{8} & G8Large (G8L)\T&9&3&128&diff. + sim. textures \\
                            & G8Medium (G8M)\TS&6&3&70&small model \\
                            & G8SmallNoR (G8SNR)\TS\B&3&3&30& xs model (no regen.)\\
         \hline
    \end{tabular}
    \caption{Multi-texture initial experiments. Sobel and Laplacian filters are used for perception. $n_h$ = number of state's hidden channels with $n_c$ communication channels and $n_g$ genomic channels. $n_f$ = number of filters for the first convolution layer of the NN.}
    \label{tab:multitex-architextures}
\end{table}

\begin{figure}[!h]
    \centering
    \includegraphics[width=350px]{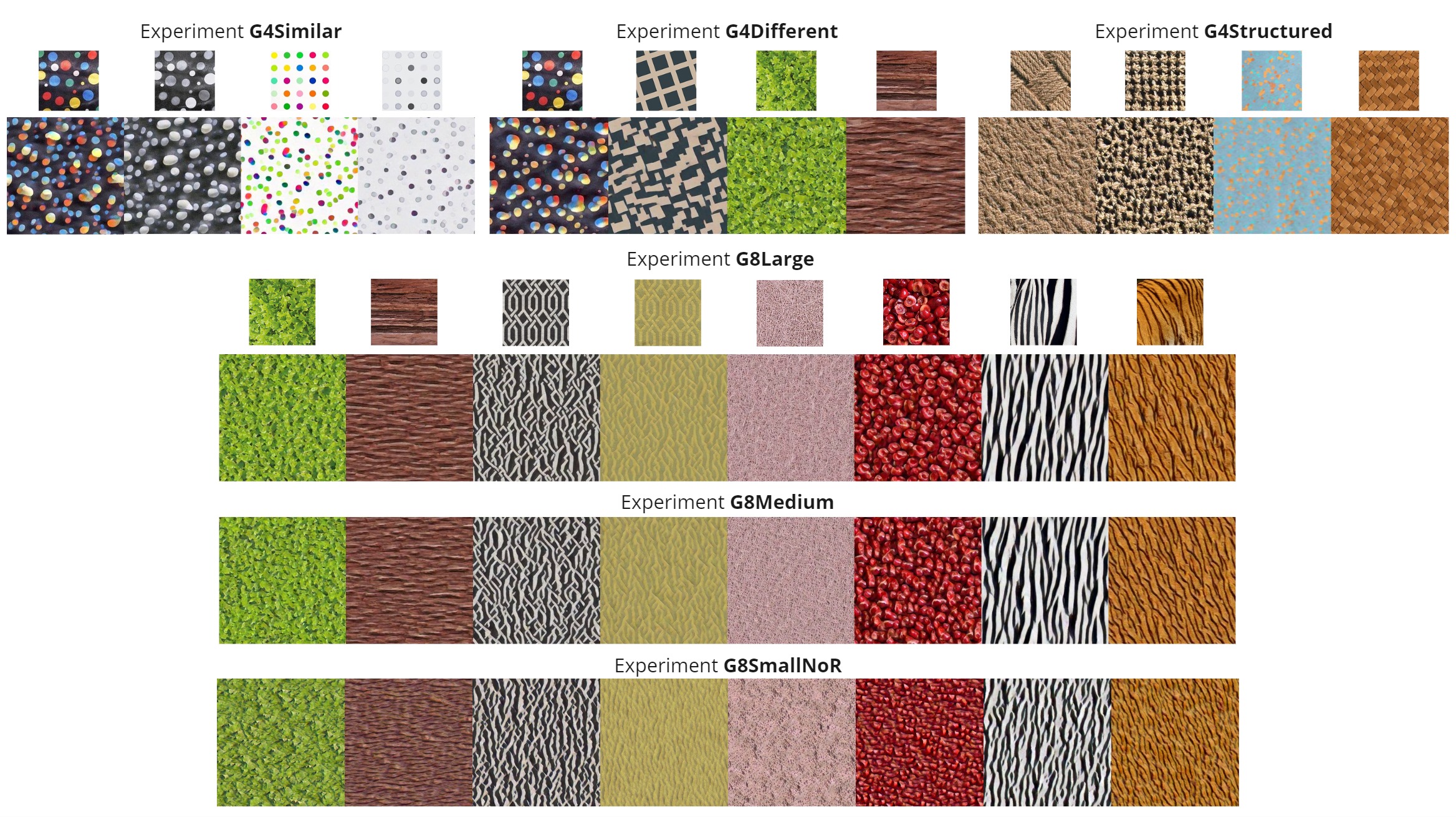}
    \caption{The results for multitexture generation, using the architectures as in \ref{tab:multitex-architextures}. Images are grouped according to architectures, each genome representing a texture. For each group of experiments, the top row represents the provided examples, and the bottom illustrates the generated textures after training.}
    \label{fig:multitex-experiments}
\end{figure}

The experiments covered various aspects in the image examples, such as similarity and regularity, and are detailed as follows: 

\begin{itemize}
     \item \textbf{Experiment G2Feasible (G2F)} tested whether a proposed genomic coding is enough for the NCA to learn to differentiate between textures. Since the results were satisfactory and the NCA learned to differentiate between the two provided textures, it opened the way for generating more textures and lead to the experiments presented below.
    \item \textbf{Experiment G4Similar (G4Sim)} follows the generation of 4 similar textures by a single NCA, enhancing shared feature representation but struggling with long-term stability of the genomes. The automaton learns to read the genome and differentiate the wanted textures, but, as seen in Figure \ref{fig:longterm-stability}, shows difficulties in maintaining the textures over a large number of iterations. This instability may be prevented by training the NCA for more epochs. All 4-genome experiments training lasted around 1h40min.

    \begin{figure}[!h]
        \centering
        \includegraphics[width=270px]{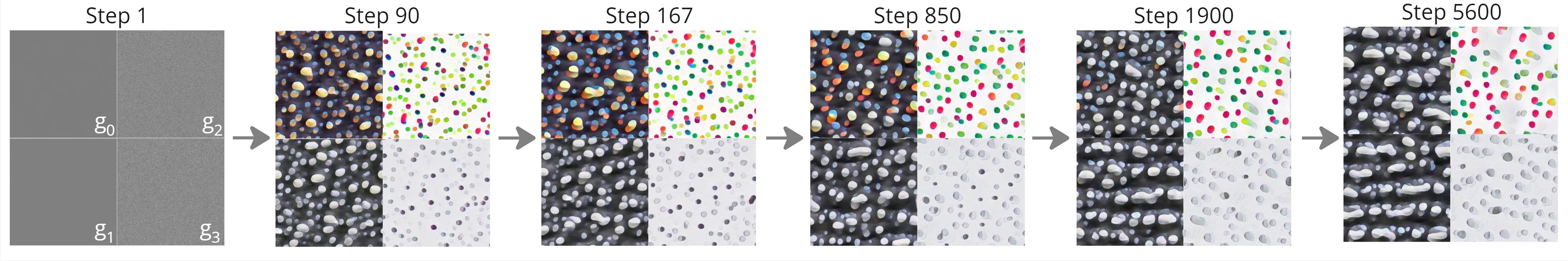}
        \caption{Genome stability for experiment G4Sim.}
        \label{fig:longterm-stability}
    \end{figure}
    
    \item \textbf{Experiment G4Different (G4Diff)} aims the generation of different textures by a single NCA, which means that different features must be interpreted and learnt. It successfully demonstrated that the automaton does not depend on shared features between textures for learning and gathers a deeper understanding of the genome and particularizes its behaviour accordingly. This experiment also uncovered an intriguing behaviour for the artificially generated texture regarding global communication along cells. Specifically, the NCA fails to achieve the global organization of the grid texture as provided for genome 2 (01), but it still produces a texture with the style specifics of the example (Figure \ref{fig:multitex-experiments}, second top). This means that the NCA learns the difference between the genomes, but fails to understand the global organization required for the highly structured texture. Referencing the G4Sim experiment, we only changed the textures we provided as examples to the automaton.
    
    \item \textbf{Experiment G4Structured (G4Str)} followed the above observation regarding the NCA behaviour on textures with large repetitive models and trained highly structured patterns that necessitate a broader communication across cells in order to test the extent of the cell's global organisation. The experiments' results hyphened a shortcoming of our approach, as the automaton can not reach the global organisation of such textures although it does imitate the style of smaller patterns (Figure \ref{fig:multitex-experiments}, third top). This limitation is tackled in Section \ref{sec:4.2LossFunctionExploration}.
    
    \item \textbf{Experiments G8Large (G8L) and G8Medium (G8M)} followed 8-texture generation on a pack of different textures, both artificially generated and real-life examples. They covered different texture categories, colors and patterns, while offering enough room for interpolation between examples. Experiment G8M was performed in order to optimize the number of parameters of the NCA used in Experiment G8L, downgrading the 10k parameter architecture to a 4270 parameter architecture with similar results, while keeping all the other training details unaltered. Training these experiments lasted approximately 1h40min.

    \item \textbf{Experiment G8SmallNoR (G8SNR)} followed 8-texture generation on the same texture pack as the other 8-genome experiments, with the scope of testing how small can our neural network be. Since G8M exhibited a slow regeneration process, we decided to drop the regeneration expectancy for this experiment. We therefore obtained a 1500 parameter NCA that can generate 8 different textures and that holds stability up until 500 steps. Roughly, this architecture size would be equivalent to 8 NCAs with 187 parameters that correctly generate the expected textures, a hard-to-complete task. However, our approach's disadvantage in this case would be the long-term instability of the genomes. This architecture trained for 1h20min.
\end{itemize}

In Figure \ref{fig:multitex-experiments} we visualize the images generated by our experiments, in groups. The conducted experiments were successful. The automaton learned to evolve numerous textures according to the planted internal signals. Most training experiments were done on an unnecessarily large neural network, as seen in comparing the results between architecture G8L of 10k parameters and architecture G8M of 4k parameters with little impact on output texture quality, visible in the second and third rows of Figure \ref{fig:multitex-experiments}. However, the smaller architecture's regeneration process is slowed down. For example, the G8L architecture visually regenerates the first genome in approximately 180 steps, whereas the G8M architecture restores it in approximately 420 steps. Another experiment with an even smaller architecture followed, with 3300 parameters ($n_c=5$, $n_f=60$), and the 8 textures was successfully developed similarly to the ones presented above, but the regeneration process lasted 700 steps and the automaton did not recover perfectly. Moreover, the genomes get corrupted around the timestamp 1000.

The images selected in Figure \ref{fig:multitex-experiments} are at timestamp 110. At inference, cellular automata can run an indefinite number of steps. In most of our experiments, we observed that the automaton held stability up until 6000 iterations (some even longer) when one genome would get corrupted and start evolving patches of some other genomes. 

Furthermore, we test the NCA's ability to generalize learnt features by analyzing its performance on interpolation tasks. Interpolation is a niche texture generation area \cite{textureInterpProbing}, of interest in computer graphics \cite{texMixing, patchBInterp}. This ability is specific to our NCA architecture, as other NCAs developed for texture synthesis cannot generate multiple textures using one single model. 

\begin{figure}[!h]
    \centering
    \includegraphics[width=250px]{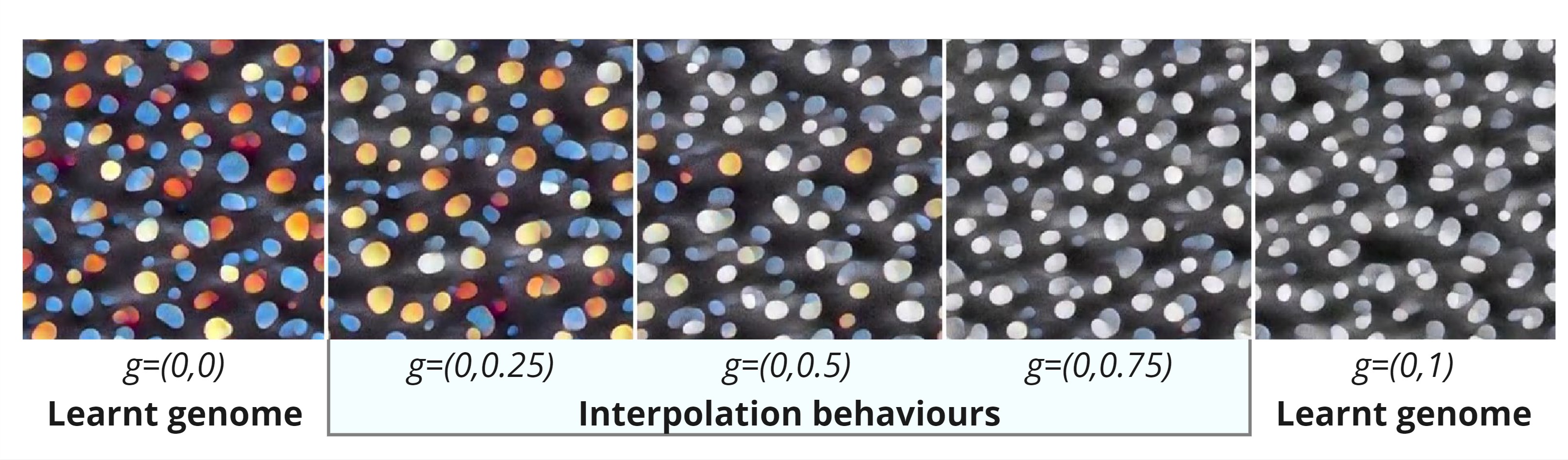}
    \caption{Interpolation of learnt genomes for G4Sim experiment.}
    \label{fig:interpolation-extended}
\end{figure}

Figure \ref{fig:interpolation-extended} illustrates the interpolation between two learnt genomes, for experiment G4Sim. We aimed to achieve a blended polka-dot texture between a coloured and grayscale version of the same texture. This is an ideal scenario, as the textures are highly similar. We can observe in Figure \ref{fig:interpolation-extended} the quality of the interpolated textures, which closely resemble those that a programmer could manually create by using color dots derived from each texture. The interpolation method avoids combining multiple colors into a single dot and ensures that the dots maintain alignment with the properties of the original genomes. This observation aligns with the statement in \cite{mordvintsev:Self-organisingTex} that the cells find an algorithm that generates the patterns. Moreover, specifically on these textures, this example can be extended into multiple color spectrums and enjoy the advantages of interpolation in creating new, blended, images.

\begin{figure}[!h]
    \centering
    \includegraphics[width=250px]{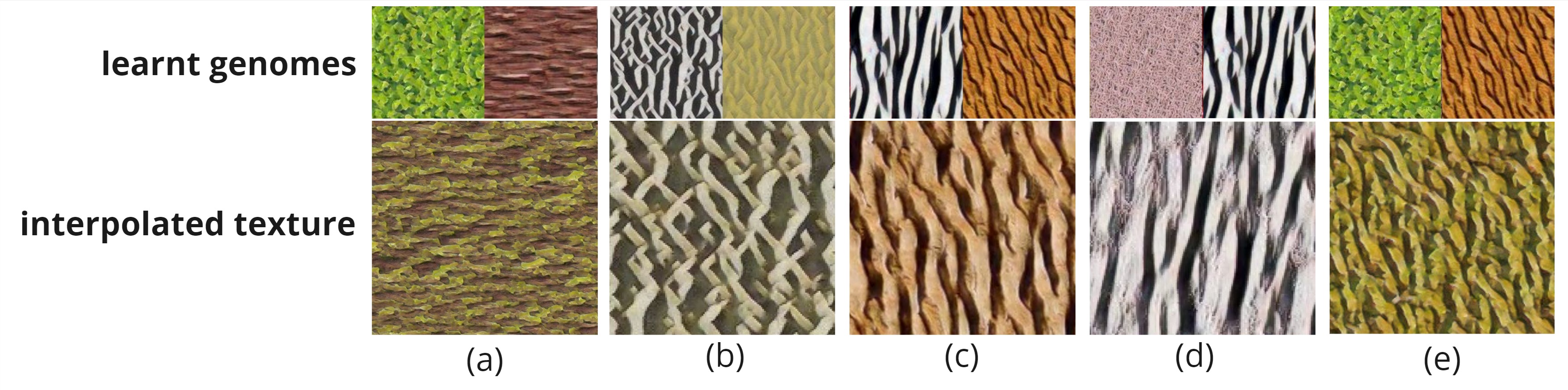}
    \caption{More interpolated textures, all for the 8-genome experiment G8L. Genomes initialized with: (a) $g=(0,0,0.5)$ (b) $g=(0,1,0.5)$ (c) $g=(1,1,0.5)$ (d) $g=(1,0.5,0)$, (e) $g=(0.35, 0.35, 0.35)$.}
    \label{fig:interpolation-more-examples}
\end{figure}

Examples covering the interpolation of more distinctive textures are represented in Figure \ref{fig:interpolation-more-examples}. These extend on the aforementioned ideas and demonstrate the quality and utility of interpolation, paving a new area of development for NCAs in texture generation. The loss function heavily influences the interpolation behaviour, and employing other loss functions will lead to varying results, as seen in the experiments conducted in Section \ref{sec:4.2LossFunctionExploration} and in \cite{gotex:65kparam}.

\subsubsection{Regeneration and grafting results} 
In addition to the experiments described in the preceding sections, some experiments related to regeneration and grafting have also been performed. These highlight the proposed architecture's extensibility while also providing a stable baseline for texture synthesis using small models.

\textbf{Regeneration} experiments were carried out to test whether our automaton can consistently inpaint a damaged region. This behaviour is studied in different NCA architectures \cite{mordvintsev:Self-organisingTex,mordvintsev2020growing,Stovold:NCARespondSignals} and the inpainting task is also covered by other texture synthesis architectures \cite{gotex:65kparam}. The outcome of our experiments emphasizes the rich representation of the genomic texture that leads to successful regeneration. All of our NCAs trained with the adapted methodology enhanced this behaviour. The NCA of experiment G8M had a slower regeneration process due to its reduced architecture.

\begin{figure}[!h]
    \centering
    \includegraphics[width=250px]{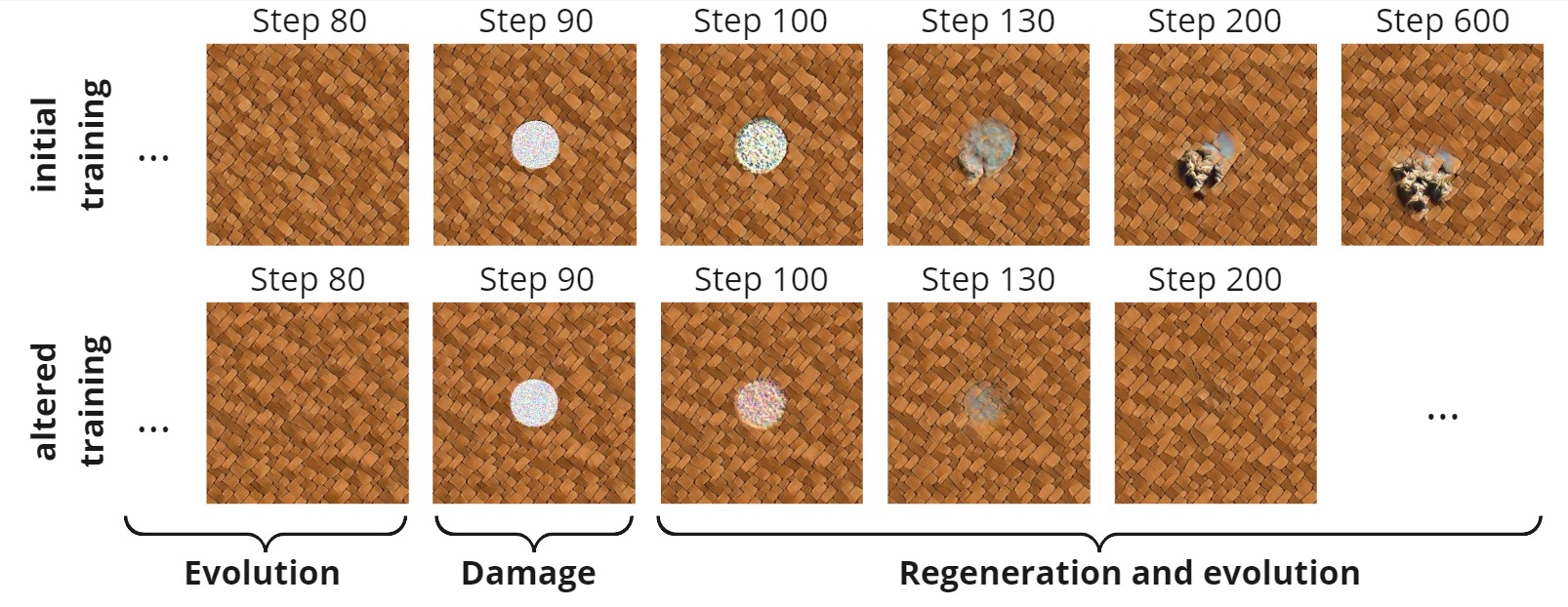}
    \caption{Regeneration before and after adjusting the training methodology, exemplified on the automaton of experiment G4Dif.}
    \label{fig:regeneration}
\end{figure}

Figure \ref{fig:regeneration} displays the NCA regeneration behaviour before and after adjusting the training methodology, as presented in Algorithm \ref{alg:regen-mtg} of Section \ref{sec:2.3OtherExp}.

\textbf{Grafting} experiments highlight the rich context in which textures can be generated and combined. As discussed, interpolation tasks involve blending between existing textures to create a new one, combining features of both in a smooth transition. On the other hand, grafting combines distinct textures onto a single surface and often requires careful alignment and blending for a cohesive appearance. Recent implementations of grafting techniques in this context include utilizing compatible neural cellular automata instances. For example, \cite{pajouheshgar2024meshNCA} defines compatible NCA instances as instances where their weights at training are initialized with those from a common trained ancestor. These instances provide the base for texture grafting, alongside interpolation at the border between the two textures provided by passing one image through both NCA instances and masking them to create the final result. While our architecture allows for this grafting type, we see the benefit of being able to obtain grafting based just on the genome channel with a single automaton.

We \textit{graft} textures by either initializing, at timestamp 0, one NCA with different genomes in the patches we want, or running a NCA on one genome, selecting a patch at a given timestamp and transferring that patch over to another NCA, with different genomes. Of course, given that the combined automaton will continue to evolve iteratively, the patches will modify shape, location and area over time. If this is an unwanted behaviour, we could create the illusion of grafting by running 2 or more instances of the NCA, each with an expected genome, and layering the generated textures over each other by masking unwanted patches for each.

\begin{figure}[h!]
    \centering
    \includegraphics[width=0.9\linewidth]{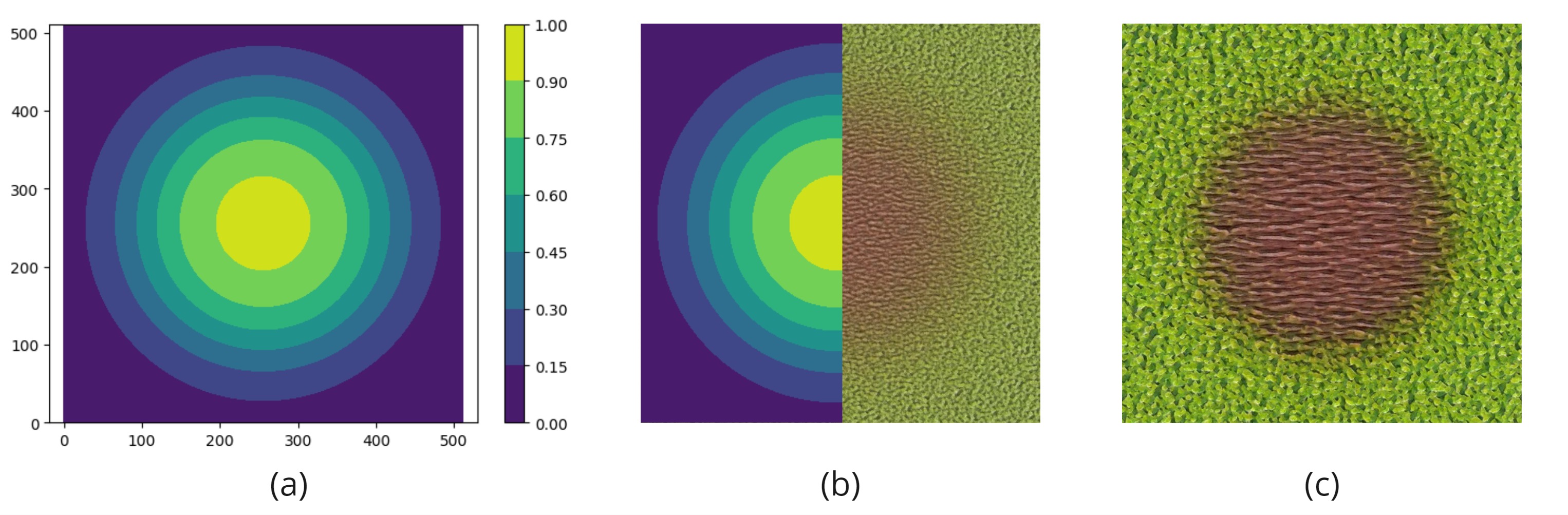}
    \caption{Grafting the G8L nca. Initialization example with genomes $g_0 = (0,0,0)$, $g_1 = (0,0,1)$. (a) Initialisation of the last genome channel, (b) Comparison of the channel initialization with the  evolution of the nca at timestamp 30, (c) The state of the nca at timespamp 110.}
    \label{fig:053_Grefare}
\end{figure}

In Figure \ref{fig:053_Grefare} a visual example of the initialization with different genomes for grafting is presented. In Figure \ref{fig:053_Grefare} (a) the yellow color represents genome $g_1 = (0,0,1)$, while the blue color represents genome $g_0 = (0,0,0)$. The shades between yellow and blue represent genomes for which the last channel has intermediate values in the range of $[0, 1]$, enabling a smooth transition between the two textures. Figure Figure \ref{fig:053_Grefare} (b) illustrates the comparison of the initialization mask with the results of the evolution of the nca at timestamp 30. The results of this initialization on the evolution of the nca at timestamp 110 can be seen in Figure \ref{fig:053_Grefare} (c).

\begin{figure}[!h]
    \centering
    \includegraphics[width=0.9\textwidth]{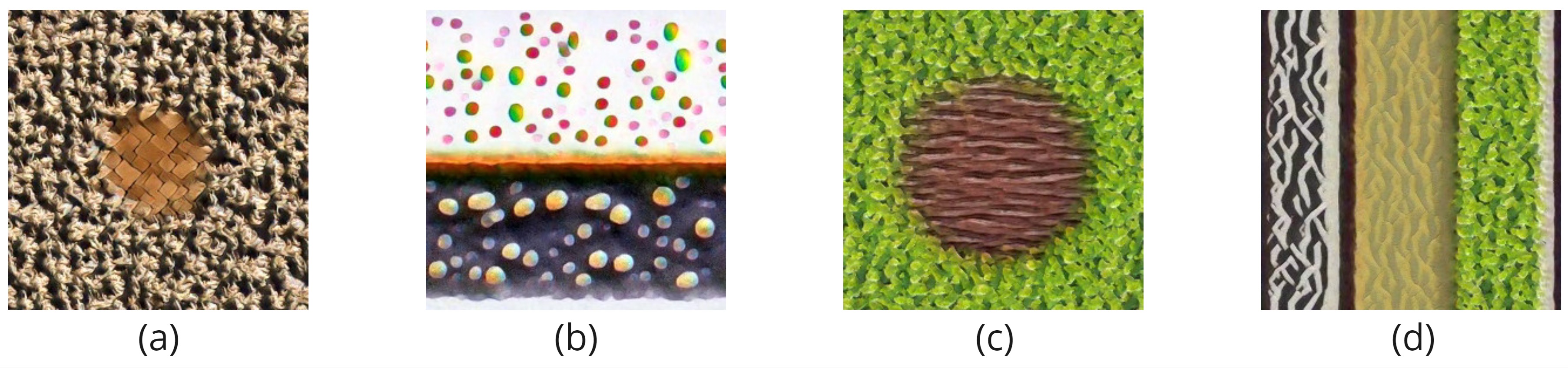}
    \caption{Grafted textures, from left to right, for NCAs of experiment (a) G4Structured: illustrates a NCA formed of cells of genome 2 with a disc of genome 1 situated in the center, (b) G4Similar: the top half is formed out of genome 3 cells, whereas the bottom half is formed out of genome 1 cells, (c) G8Large: cells situated similarly to (a), of genome 1 and 2, (d) G8Large: cells situated in three equally distributed vertical stripes of genomes 4, 3 and 1.}
    \label{fig:grafting}
\end{figure}

Figure \ref{fig:grafting} illustrates a four textures generated through grafting techniques. We observe that the automaton can develop patches of specific genomes. Moreover, we see a communion where the two or more genomes collide, forming a consistent boundary transition area of cells, whether visualized as a barrier (as seen in (b), (d)) or a smooth transition ((a), (c)). Nevertheless, the transition behavior between genomes remains consistent across the intersection line.

In the light of all the described experiments, two main questions arose. Firstly we were intrigued by the role of the generation and stability of the desired texture. Thus we studied if the automaton preserves the genome information in the texture cell during evolution and by this learns to discern between the different textures. The second question was about the influence of the loss function on improving the generation of highly-structured textures that require broad-image communication. The results obtained using two different loss functions were compared. In the following these two aspects are presented in more detail. 

\subsection{Preservation of the genome}
\label{sec:preserveGenome}

Given the diversity of the generated textures in our experiments, it is essential to investigate whether the automaton preserves and generates textures based solely on the perception stage and vicinity properties, or if it also maintains genome information to ensure stability.

\begin{figure}[!h]
    \centering
    \includegraphics[width=100px]{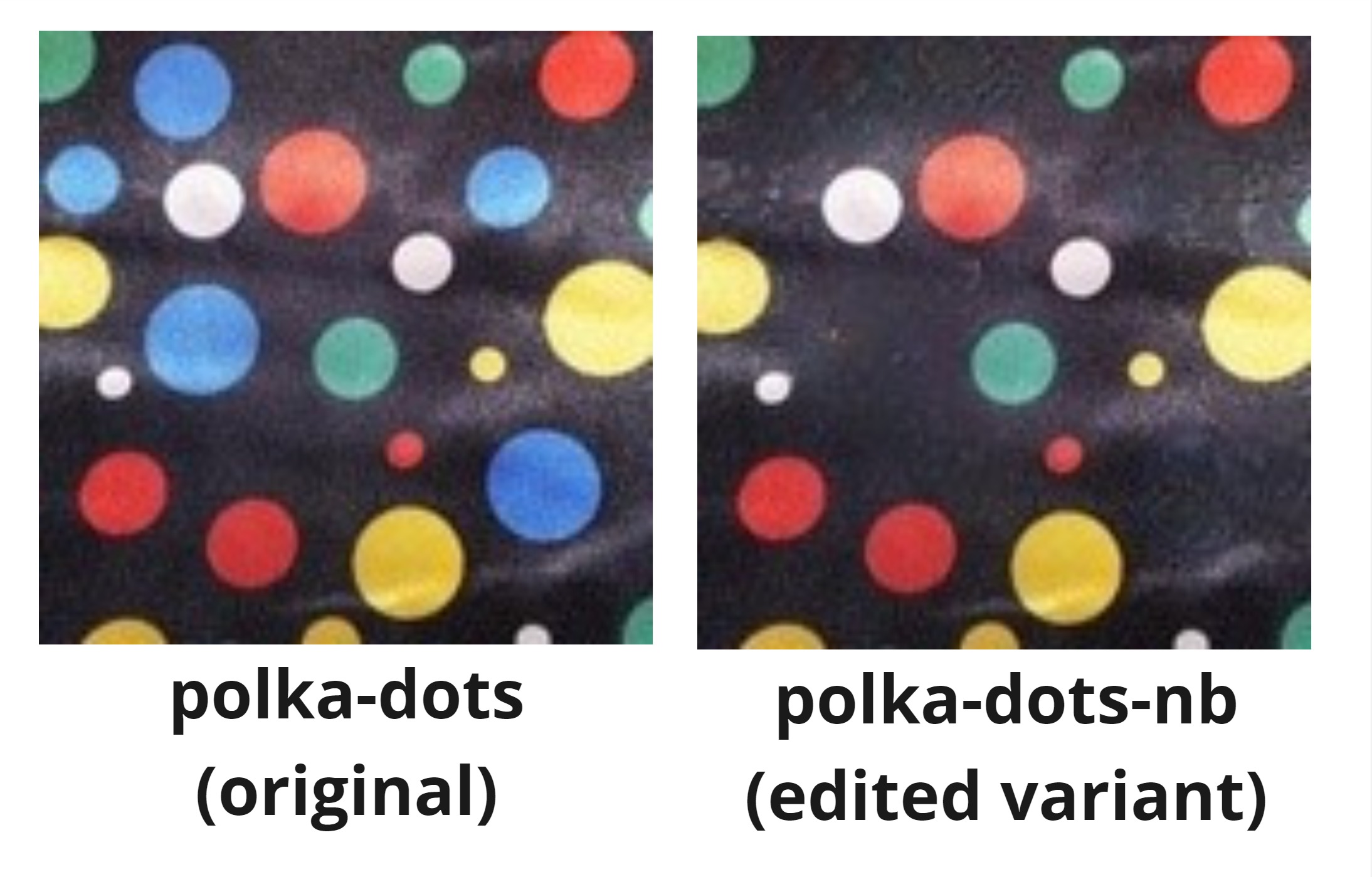}
    \caption{The 2 textures used for this experiment, the left one corresponds to the genome 0, the right one - genome 1.}
    \label{fig:polka-edited}
\end{figure}

To investigate this issue, we based our approach on experiment G2F, where the two genomes correspond to the same texture in both color and grayscale. The automaton is more unstable given the similarity of the textures, but we cannot attribute this to the vicinity similarity, given the different colours. Therefore, we create two similar textures and test whether the automaton can learn the difference between them and remain stable over time. We selected the \code{dotted\_0201} texture \cite{db:Odtd} (Figure \ref{fig:polka-edited} left) and created a similar one by deleting the blue polka-dots (Figure \ref{fig:polka-edited} right). Note that other colours were not modified, only the blue dots were deleted. 

We name this experiment \textbf{G2Preserve (G2Psv)}. The state vector of the NCA for these two textures has $n_s=13$ values, representing the 3 RGB channels, $n_c=9$ communication channels and $n_g=1$ genome channel. The genome channel is set to 0 for the first texture and to 1 for the one without blue dots. We employ $n_f=128$ filters for the first convolutional layer and train the NCA for 10000 epochs. The training step took 1h30min.

\begin{figure}[!h]
    \centering
    \includegraphics[width=300px]{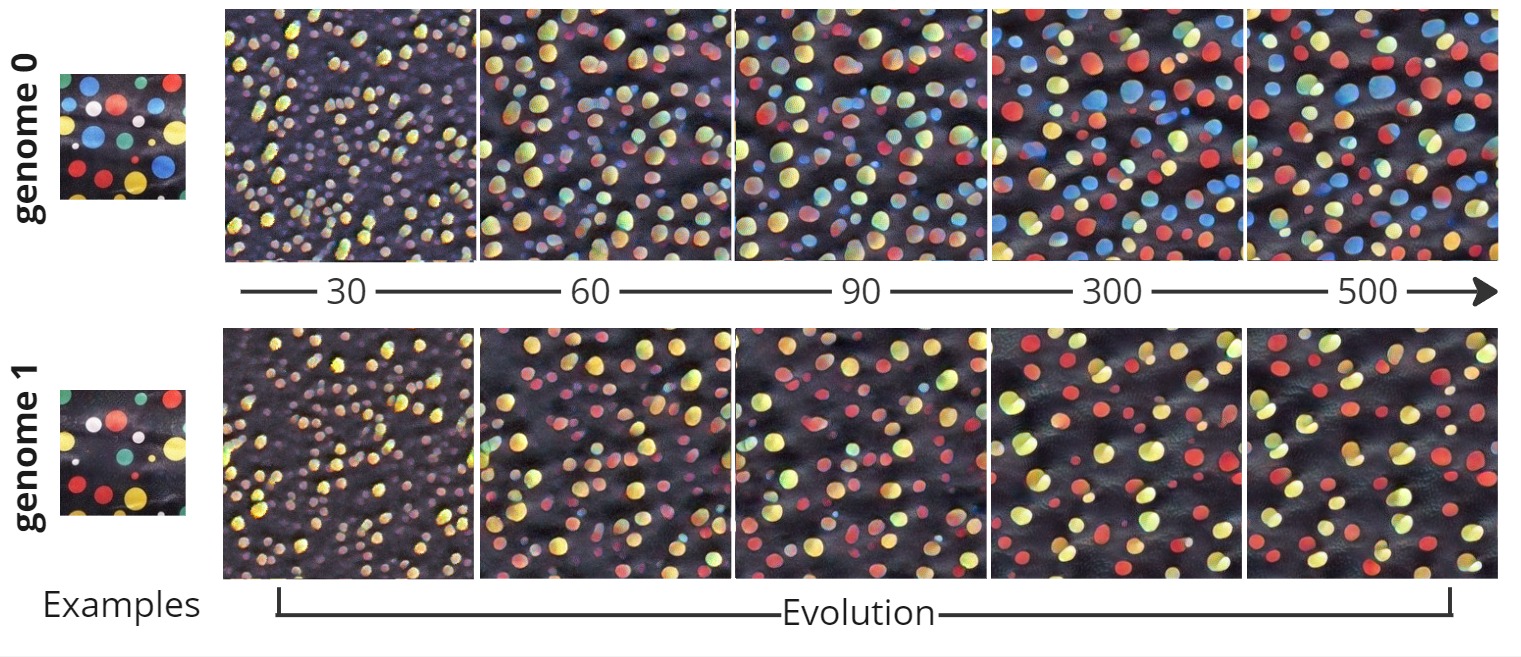}
    \caption{The evolution of the automaton based on the selected genome, in the first row - genome 0, in the second row - genome 1. The examples on the first column of size $128 \times 128$ are provided for comparison to the generation of the textures of size $256 \times 256$.}
    \label{fig:vicinity-evolution}
\end{figure}

Figure \ref{fig:vicinity-evolution} displays the texture generation results for each of the two genomes after different timestamps. It can be observed that the automaton has learned to differentiate between the two textures despite their extreme similarity, suggesting that the automaton considers the genome throughout evolution. 

We also remark an intriguing, unexpected behaviour. Monitoring the evolution of the NCA, we notice in the second texture (with genome 1) the emergence of small blew dots at evolution step 60. While not yet prominent, it would have likely continued evolving them into dots akin to those seen in the texture generated by genome 0, had the automaton not considered the genome (see this transformation of blue dots throughout genome 0 generation in Figure \ref{fig:vicinity-evolution} steps 60-90). However, for genome 1 the formed dots are dimmed and at step 500 no such dots are in formation, supporting the idea that the automaton does keep the genome information throughout evolution.

\begin{figure}[!h]
    \centering
    \includegraphics[width=350px]{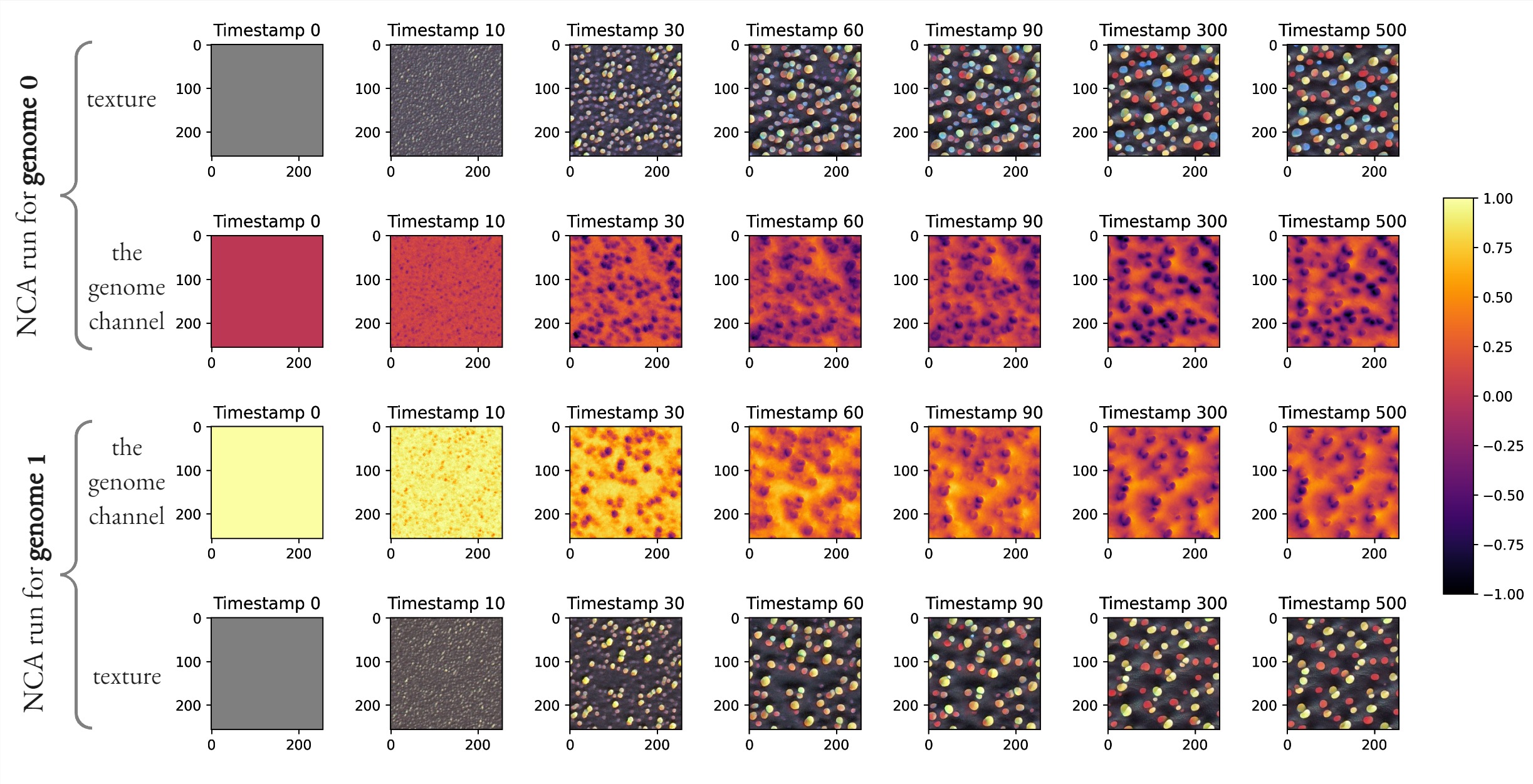}
    \caption{Analysis of the genome channel for the G2Psv NCA. The first and bottom rows illustrate the generated textures, whereas the second and third rows represent the corresponding genome channels.}
    \label{fig:genome-similar-preservation}
\end{figure}

To further support this hypothesis, we provide two figures constructed in similar manner. Figure \ref{fig:genome-similar-preservation} illustrates the generated textures alongside the genome channel values for two inferences of this experiment, one with cells' belonging to genome 0 (top two rows) and one with cells' belonging to genome 1 (bottom rows). The genome channel is visualized through the colormap corresponding to the colorbar represented on the right of the image, where black corresponds to a genome channel value of -1, and yellow to a genome channel of 1. We observe that the genome channel values are as expected up until around timestamp 60, for genome 0, the channel's values are close to 0 and for genome 1 the channel's values are close to 1. Although the values converge at later timestamps, we pose that the automaton has assigned certain genomic values to each texture style specifics and uses them, alongside the other channels, to maintain texture stability over time. To support this statement, we direct the attention towards the genome channel for the genome 0 NCA run, at timestamp 500 (last column of row 1 in Figure \ref{fig:genome-similar-preservation}). The lower values, approaching -1, visualized as black and dark-purple dots correspond directly to blue dots in the RGB correspondent of the NCA state (row 0, last column), meaning, the automaton has attributed low genome values to texture features specific to the provided example for genome 0, and higher values otherwise, as the common features represent both genome 0 and 1. For the genome 1 generation (third row) we see higher values overall, no deep purple spots in the genome channel at timestamp 500. The lower genome values correspond to the blue, faded dots and areas that are prone to the generation of such dots in the texture; values are later corrected by the self-organising system.

We also provide the last genome channel analysis for textures generated by experiment G2Diff. The textures representing genome 00 and 01 have no similar features, as the ones discussed in the aforementioned example, resulting in a clearer divide between the values representing the last genome channel. The last genome channel values for the top rows keep values around 0, as the values for the generation of genome 01 keep higher values throughout evolution.

\begin{figure}[!h]
    \centering
    \includegraphics[width=350px]{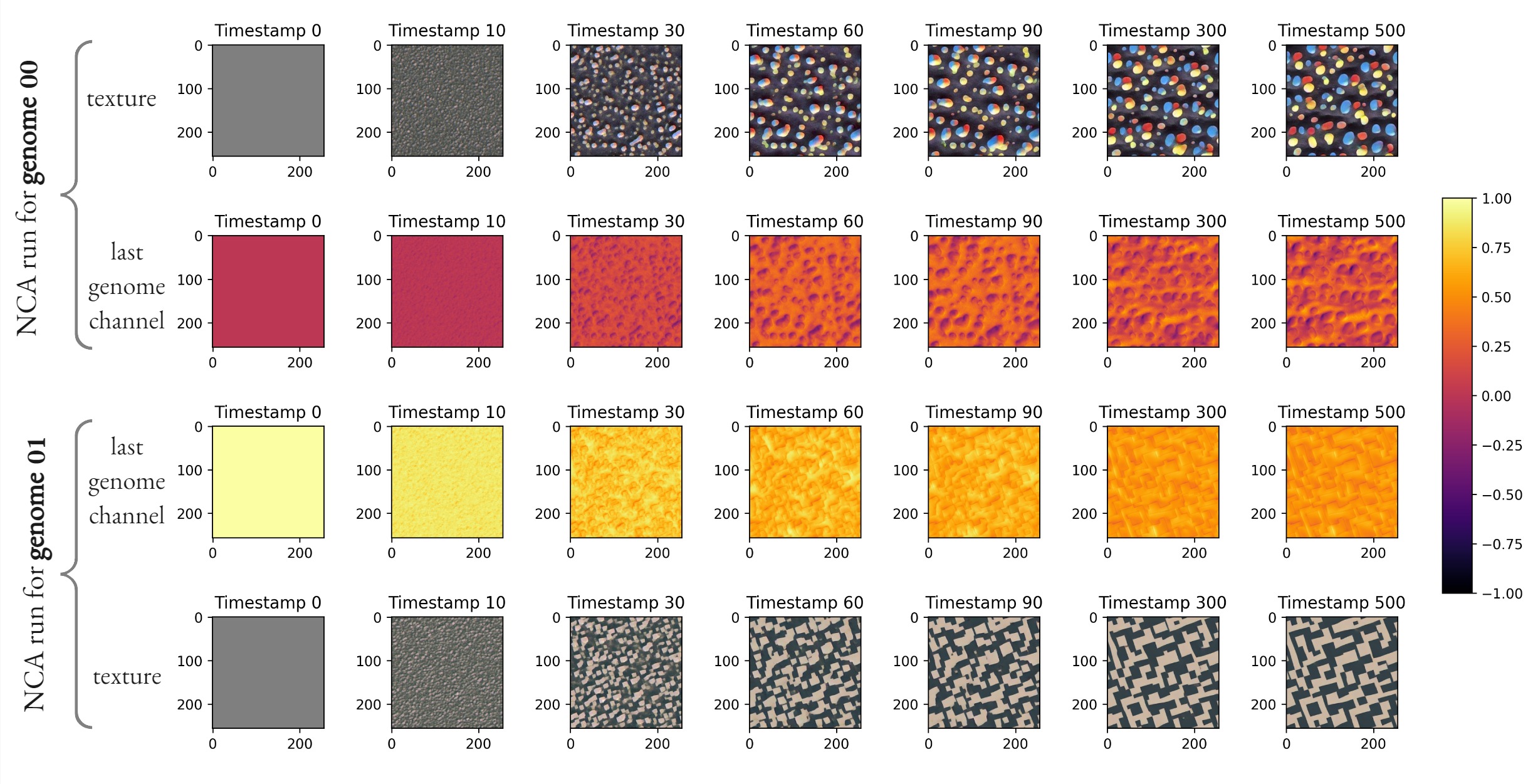}
    \caption{Analysis of the genome channel for G4Diff NCA.}
    \label{fig:genomedifferentpreservation}
\end{figure}

Furthermore, upon comparing the textures generated with automaton G2Psv for genome 0 with those generated by genome 0 for the G4Sim NCA, we remark an uptick in the occurrence of red, orange, and yellow dots, accompanied by a decrease of blue and green dots. We also observe throughout training that the G2Psv automaton learned to only generate the edited (polka-dots-nb) texture up until the 2000 epoch. Before and at this timestamp, the automaton generates the edited version of the texture no matter what genome we set for in the seed. Only after this epoch, the automaton learns to generate the blue dots corresponding to genome 0. This highlights the complex interactions between the textures during training and the NCAs tendency to generalize learnt behaviour. In these experiments, the NCA has learnt to evolve red, orange and yellow dots faster through the texture provided for genome 1, consequently influencing the generation process for genome 0.

\subsection{Loss function exploration}
\label{sec:4.2LossFunctionExploration}
\begin{figure}[!h]
    \centering
    \includegraphics[width=300px]{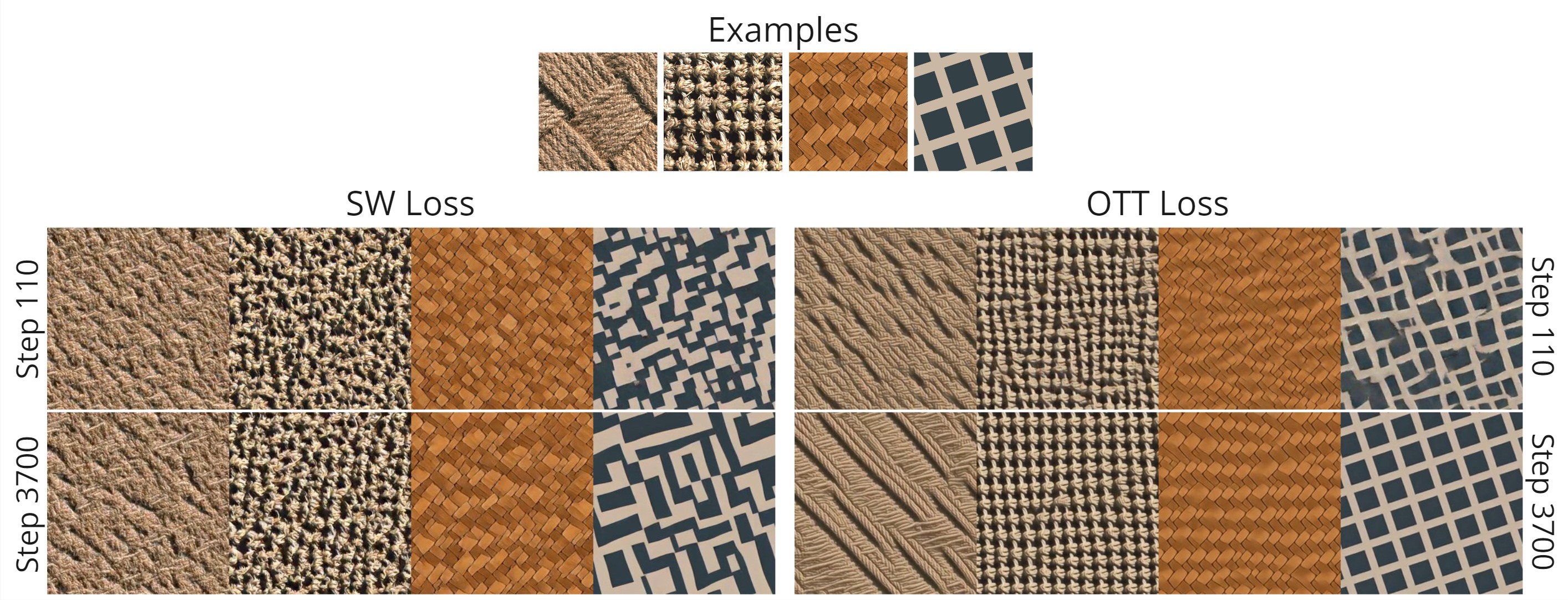}
    \caption{Generation of fabric textures (extracted from \cite{db:VisTex}) and a grid texture (extracted from \cite{db:Odtd}) using the SW Loss and OTT Loss.}
    \label{fig:broad-communication}
\end{figure}

Utilizing the SW loss function, the automaton performs strongly on both near-regular and irregular structures. However, it exhibits weaker performance on textures that necessitate broader communication across the automaton, such as images that contain large repetitive patterns. Examples of such given and generated textures are depicted in Figure \ref{fig:broad-communication}. It's noteworthy that this behavior is intentional and expected in many cases, as it allows for the capture of fine details while disregarding potential irregularities in the given examples, without straining the texture. The NCA does capture the style of the given image, but does not reach a similar global state. Nevertheless, for the showcased instances, we would prefer the NCA to prioritize learning the knitted, knotted, checkered or tiling structures over focusing solely on the finer details of the examples.

\begin{figure}[!h]
    \centering
    \includegraphics[width=300px]{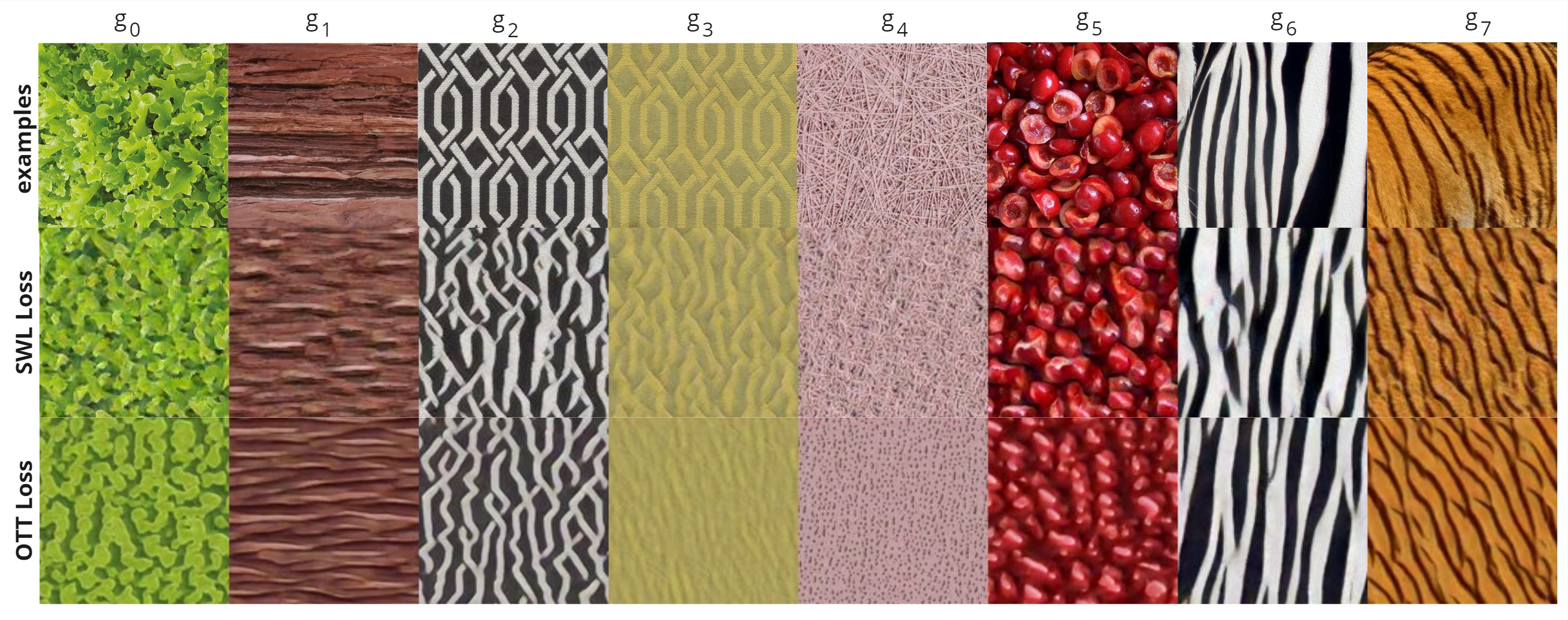}
    \caption{Based on the examples of the top row, generated textures (size $256 \times 256$) using the same architecture and parameters, but utilizing SWL Loss and, respectively, OTT Loss during training. %Both 8-genome models use 4270 parameters.
    }
    \label{fig:swl-ott-8genome}
\end{figure}

For comparison, we employ the loss function termed OTT Loss proposed by \cite{mordvintsev:uNCA}, specialized for regular patterns. We conducted a 4-genome experiment on the regular patterns of Figure \ref{fig:broad-communication} using the architecture details as presented in Section \ref{sec:2.2MultiTex} for experiment G4Sim, using our SW Loss and OTT Loss respectively. A further comparison of the performance of the two losses is obtained by the 8-genome experiment with 4270 parameters, utilizing the G8M architecture, and the outcomes can be visualized in Figure \ref{fig:swl-ott-8genome}. Both experiments highlight the limitations of each loss: SWL produces favorable results on most textures, while OTT Loss excels in generating structured patterns. Examining the generation of the fibrous texture (Figure \ref{fig:swl-ott-8genome} $g_4$), it is notable that SWL captures a better understanding of the texture, while OTT Loss generates small dots against a pink background. The interlaced texture (Figure \ref{fig:swl-ott-8genome} $g_3$), characterized by similar colours for foreground lines and background is less accurately captured by the OTT Loss, as are the properties of the frilly texture (Figure \ref{fig:swl-ott-8genome} $g_0$) and pitted texture (Figure \ref{fig:swl-ott-8genome} $g_5$). 

\begin{figure}[!h]
    \centering
    \includegraphics[width=200px]{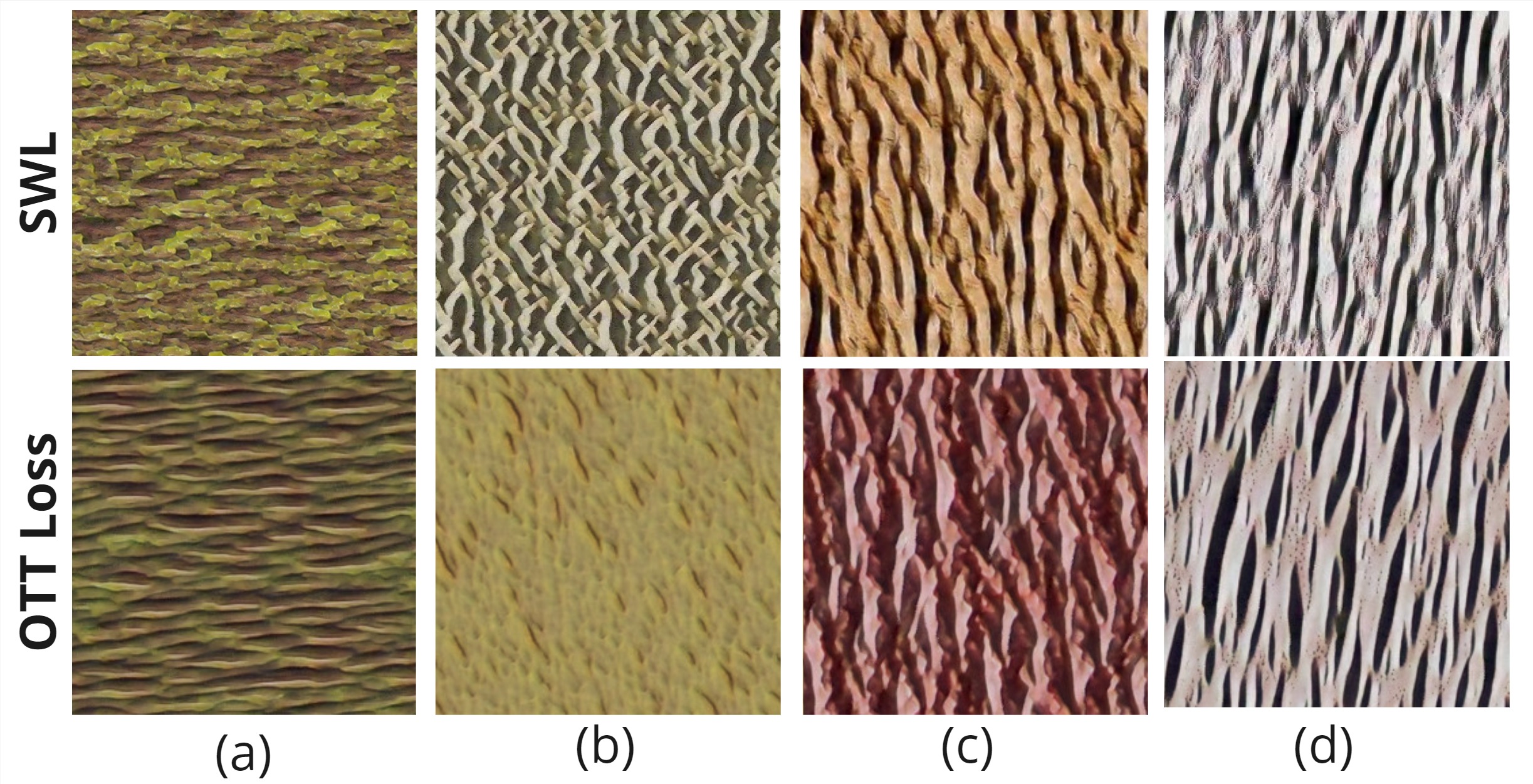}
    \caption{Interpolation behaviour %(textures of size 256x256) 
    comparison of the NCA trained on the same methodologies with 4270 parameters but with SWL and, respectively, OTT Loss. Genomes for interpolation are considered as in Figure \ref{fig:interpolation-more-examples}.}
    \label{fig:swl-ott-8genome-interp}
\end{figure}

Overall, SWL captures more details, as illustrated in Figure \ref{fig:swl-ott-8genome}, given that both trainings rely on 4270 parameters. However, the OTT Loss addresses the aforementioned shortcoming of the SWL in enhancing broader communication along the cells, as depicted in Figure \ref{fig:broad-communication}. Regeneration is slowed down using the OTT Loss (from 410 steps for regenerating the first genome to 1210), and interpolation leads to images as displayed in Figure \ref{fig:swl-ott-8genome-interp} (for comparison with the textures displayed in Figure \ref{fig:interpolation-more-examples}).

We conclude that the OTT loss covers the shortcoming of the SW Loss for the NCA, which does not accurately reproduce textures with regular, relatively large patterns. This makes OTT loss the preferred one int the case of such texture (ideally artificial \cite{mordvintsev:uNCA}) pattern generation. For a broader approach, we consider that SWL covers most cases of the presented examples. A hybrid approach may also lead to better results for both regular and irregular patterns and is one of the things to consider in future experiments.

\section{Conclusions}
\label{sec:5Conclusions}

Neural cellular automata are an active research field with many promising future opportunities. Self-organising structures are both studied from the software point of view and hardware implementations are emerging. Cell division, regeneration, and grafting offer promising prospects in the context of physics, robotics and swarm robotics, artificial intelligence and biology, offering a captivating approach to studying and understanding dynamic, self-organising systems. Real-time robust synthesis of high-quality texture pictures can be achieved using the lightweight NCA. More significantly, it exhibits an amazing zero-shot generalization capability to several post-training adjustments, including local coordinate transformation, speed control, and resizing. 
%Neural cellular automata offer an efficient approach to texture generation due to their efficient sampling, their ability to handle complexity, memory efficiency, and reusability. By leveraging these advantages, NCAs can produce high-quality, detailed textures with lower computational costs and greater adaptability compared to traditional methods. 

This study aimed at increasing the usability of neural cellular automata in the context of texture generation, pinpointing and providing a solution for the limitation of using for each texture a dedicated automaton. For this scope, we applied the idea of providing model context through internal signals, previously used in NCAs trained for growing models from one cell \cite{Stovold:NCARespondSignals}. By employing a similar approach, we developed a novel architecture that enriches one NCA to generate multiple textures. We improved the training methodology to enhance the generation and stability of all textures, by cycling through genomes in the genome replacement stage of batch preparation. Also, we provided qualitative and quantitative analysis regarding the extent to which our automaton preserves and uses the defined genome channels throughout evolution, supporting our statement that the genome channels are indeed maintained and further used in the automaton's development. In essence, we trained the automaton to express the genomically coded signals accordingly, thus having the self-organising system form and behave conclusively with its genome coding. In order to achieve satisfying results on most types of textures, we employed an efficient loss function, Sliced Wasserstein Loss, supporting the statement that the SWL better captures style than former Gram-based solutions, utilized in the base inspiration for our implementation \cite{mordvintsev:Self-organisingTex}.

Moreover, we extended the study of neural cellular automata in the context of texture interpolation, an area previously inaccessible due to the restricted ability of only generating one texture per automaton. Interpolation is natural to our automaton, as it can derive at inference the styling for an intermediate texture between two learnt examples. We studied this behaviour and analyzed its interactions with the used loss function. Also, we examined grafting techniques using a single instance of a NCA as a host for cells belonging to different genomes, as opposed to former solutions where multiple instances were used to test this behaviour. Visual results were displayed and discussed.

To conclude, this study treated neural cellular automata as a model of morphogenesis and utilized genomic coding to make it exhibit the wanted stylistic properties of textures based on given examples. We encourage the use of NCA in software solutions by providing an extensible yet compact and easy-to-train architecture for texture synthesis. The proposed model was also studied in adjacent research contexts, such as regeneration, grafting and interpolation, and exhibits promising abilities, encouraging future developments of embedding expected behaviour into the NCA model. Furthermore, given the interest in extending NCAs to 3D textures, our study can also contribute to offering new opportunities for improvement and generalization in this area.

\bibliographystyle{unsrt}
\bibliography{bibliography}
%\printbibliography

\end{document}